\PassOptionsToPackage{square,sort,comma,numbers}{natbib}
\documentclass{article}

\usepackage{amsmath,amsfonts}
\usepackage{amssymb,amsopn}
\usepackage{bm} % bold symbol

\usepackage{graphicx}  %Required

\newcommand{\mbf}[1]{\mathbf{#1}} % vector, matrix
\newcommand{\bs}[1]{\boldsymbol{#1}} % boldsymbol

\newcommand{\sigm}{\operatorname{sgm}}

\newcommand{\kl}{\operatorname{KL}}

\newcommand{\rnn}{\operatorname{RNN}}
\newcommand{\softmax}{\operatorname{Softmax}}
\newcommand{\diag}{\operatorname{diag}}

\def\D{\mathcal{D}}
\def\loss{\mathcal{L}}

\def\T{\mathsf{T}}

\newlength{\widebarargwidth}
\newlength{\widebarargheight}
\newlength{\widebarargdepth}

\newcommand{\eat}[1]{}

\newcommand{\bx}{\mathbf{x}}
\newcommand{\bX}{\mathbf{X}}
\newcommand{\bbX}{\mathcal{X}}

\newcommand{\bbZ}{\mathcal{Z}}
\newcommand{\bz}{\mathbf{z}}
\newcommand{\bd}{\mathbf{d}}

\newcommand{\N}{\mathcal{N}}

\newcommand{\by}{\mathbf{y}}
\newcommand{\bY}{\mathbf{Y}}
\newcommand{\bZ}{\mathbf{Z}}
\newcommand{\bM}{\mathbf{M}}
\newcommand{\bh}{\mathbf{h}}

\newcommand{\ba}{\mathbf{a}}
\newcommand{\bW}{\mathbf{W}}
\newcommand{\bb}{\mathbf{b}}
\newcommand{\bI}{\mathbf{I}}

\newcommand{\E}{\mathbb{E}}
\newcommand{\real}{\mathbb{R}}

\newcommand{\bv}{\mathbf{v}} 
 
\newcommand{\bg}{\mathbf{g}} 
\newcommand{\bw}{\mathbf{w}} 
\newcommand{\bc}{\mathbf{c}}
\newcommand{\bu}{\mathbf{u}}
 
\newcommand{\bbeta}{{\bs\beta}} 

\usepackage[preprint]{nips_2018}

\usepackage[utf8]{inputenc} % allow utf-8 input
\usepackage[T1]{fontenc}    % use 8-bit T1 fonts
\usepackage{hyperref}       % hyperlinks
\usepackage{url}            % simple URL typesetting
\usepackage{booktabs}       % professional-quality tables
\usepackage{amsfonts}       % blackboard math symbols
\usepackage{nicefrac}       % compact symbols for 1/2, etc.
\usepackage{microtype}      % microtypography
\usepackage{graphicx}
\usepackage{subfigure}
\usepackage{natbib}
\usepackage{hyperref}

\usepackage{multirow}
\usepackage{wrapfig}

\usepackage{textcomp}
\usepackage{cals}
\usepackage[table,xcdraw]{xcolor}

\newcommand{\hb}[1]{{\color{blue}{\small\bf\sf [Haebeom: #1]}}}

\title{Uncertainty-Aware Attention for\\ Reliable Interpretation and Prediction}

\author{Jay Heo$^{1,3}$\thanks{Equal contribution}\ , Hae Beom Lee$^{1,3}$\footnotemark[1]\ , Saehoon Kim$^3$, Juho Lee$^{3,5}$,\\ \bf Kwang Joon Kim$^4$, \bf Eunho Yang$^{2,3}$, \bf Sung Ju Hwang$^{2,3}$\\
UNIST$^1$, Ulsan, South Korea, KAIST$^2$, Daejeon, South Korea, \\ AITrics$^3$, Yonsei University College of Medicine$^4$, Seoul, South Korea, \\ University of Oxford$^5$, Oxford, England\\
\texttt{$\{$jayheo7, hblee$\}$@unist.ac.kr, $\{$sjhwang82, eunhoy$\}$@kaist.ac.kr}\\
\texttt{shkim@aitrics.com, preppie@yuhs.ac, juho.lee@stats.ox.ac.uk}}

\begin{document}
% \nipsfinalcopy is no longer used

\maketitle

\begin{abstract}
Attention mechanism is effective in both focusing the deep learning models on relevant features and interpreting them. However, attentions may be unreliable since the networks that generate them are often trained in a weakly-supervised manner. To overcome this limitation, we introduce the notion of \emph{input-dependent uncertainty} to the attention mechanism, such that it generates attention for each feature with varying degrees of noise based on the given input, to learn larger variance on instances it is uncertain about. We learn this \emph{Uncertainty-aware Attention (UA)} mechanism using variational inference, and validate it on various risk prediction tasks from electronic health records on which our model significantly outperforms existing attention models. The analysis of the learned attentions shows that our model generates attentions that comply with clinicians' interpretation, and provide richer interpretation via learned variance. Further evaluation of both the accuracy of the uncertainty calibration and the prediction performance with ``I don't know'' decision show that UA yields networks with high reliability as well.
\end{abstract}

\section{Introduction}
%Despite its recent success, deep learning is not yet fully ready to be actively applied to mission-critical tasks such as healthcare. 
For many real-world safety-critical tasks, achieving high reliablity may be the most important objective when learning predictive models for them, since incorrect predictions could potentially lead to severe consequences. For instance, failure to correctly predict the sepsis risk of a patient in ICU may cost his/her life. Deep learning models, while having achieved impressive performances on multitudes of real-world tasks such as visual recognition~\cite{alexnet,resnet}, machine translation~\cite{Bahdanau15} and risk prediction for healthcare~\cite{retain,futoma17}, may be still susceptible to such critical mistakes since most do not have any notion of predictive uncertainty, often leading to overconfident models~\cite{guo17,deep_ensembles} that are prone to making mistakes. Even worse, they are very difficult to analyze, due to multiple layers of non-linear transformations that involves large number of parameters. 

\begin{figure*}[ht!]
\vspace{-0.1in}
\begin{center}
\small
\begin{tabular}{c c c}
\includegraphics[width=2.8cm]{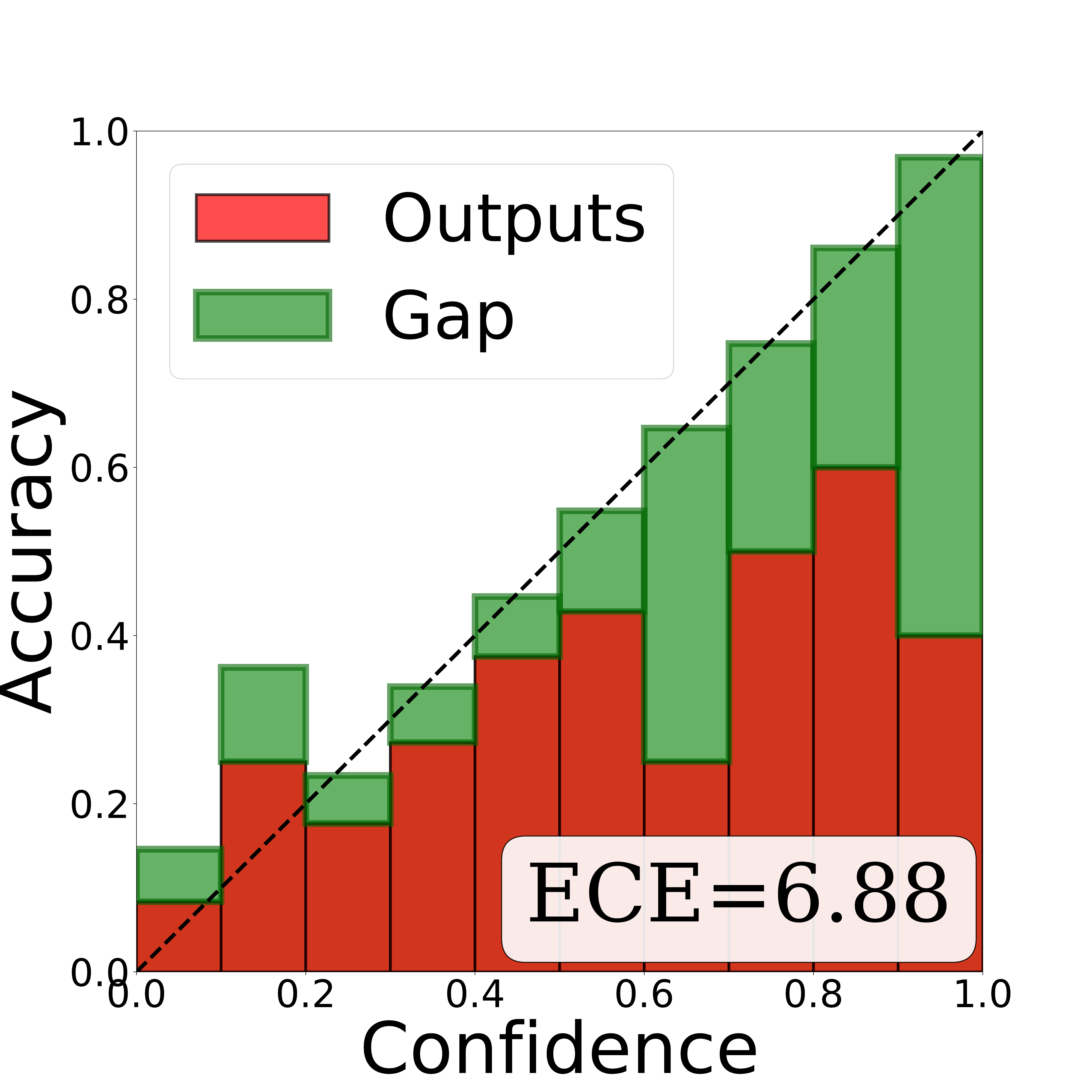}&
\hspace{0.3in}
\includegraphics[width=2.8cm]{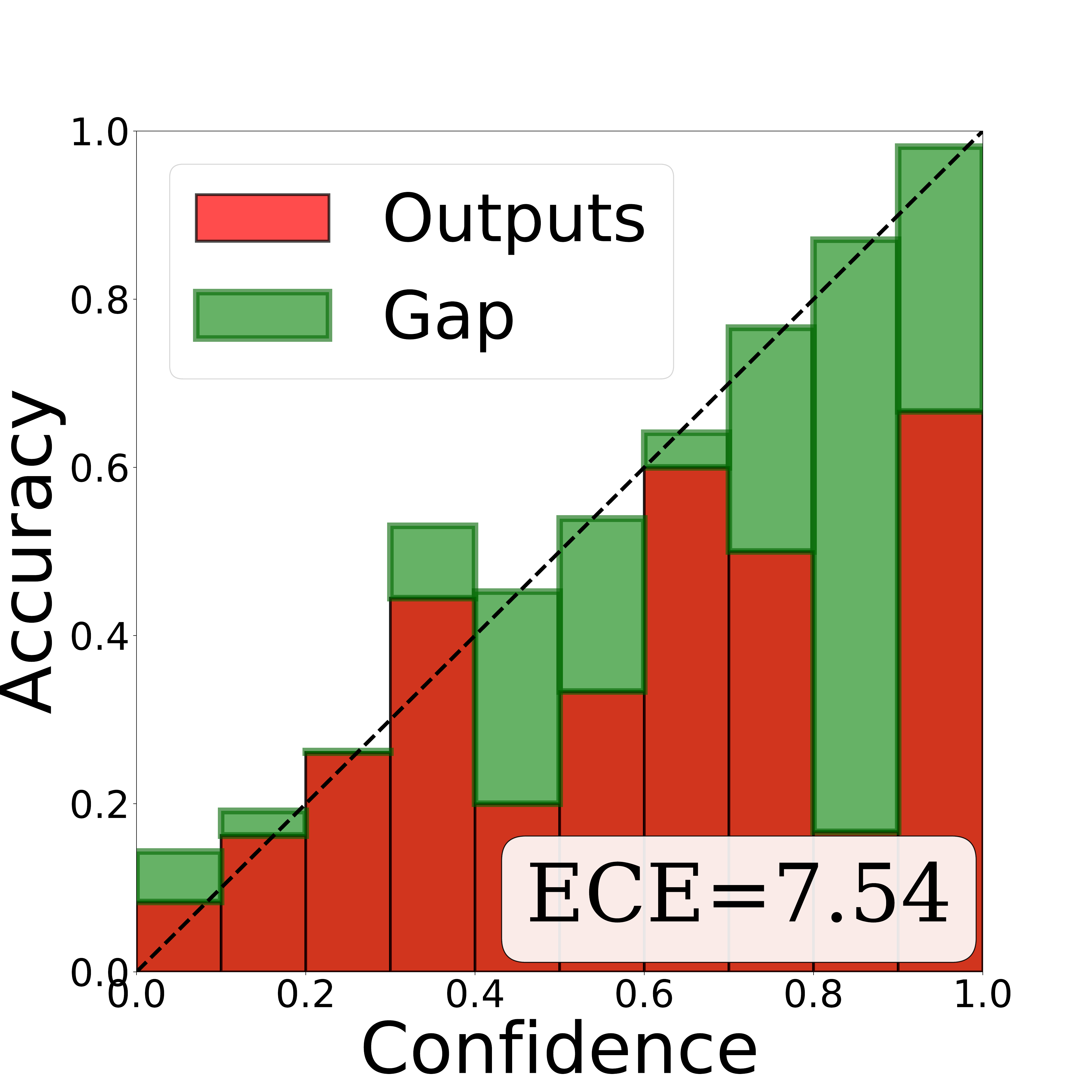}&
\hspace{0.3in}
\includegraphics[width=2.8cm]{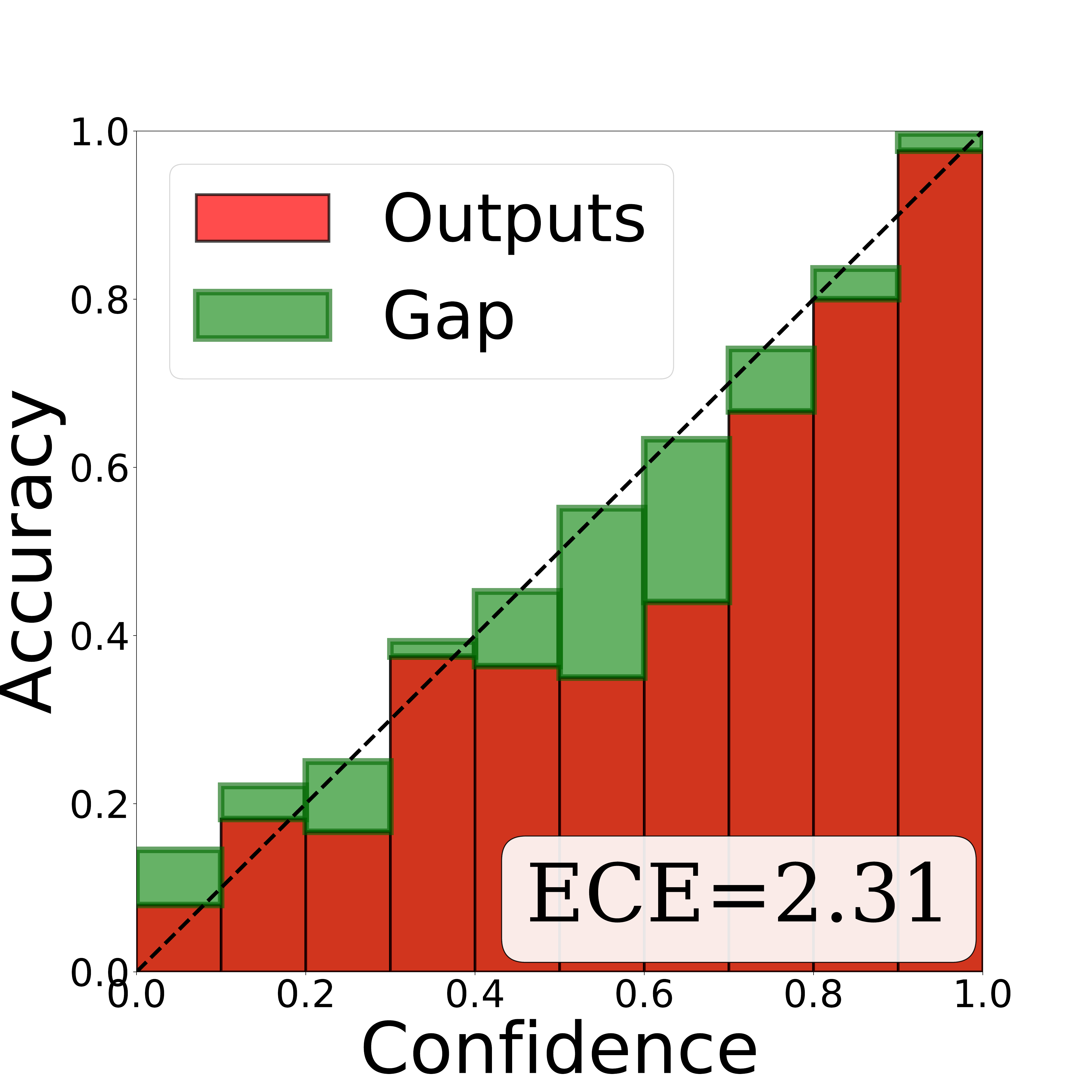}\\
(a) Deterministic Attention~\cite{retain} & (b) Stochastic Attention~\cite{show_attend_tell} & (c) Uncertainty-aware Attention (Ours) \\
\end{tabular}
\end{center}
\vspace{-0.1in}
\caption{\small Reliability diagrams~\cite{guo17} which shows the accuracy as a function of model confidence, generated from RNNs trained for mortality risk analysis from ICU records (PhysioNet-Mortality). ECE~\cite{ece} in \eqref{eq:ece} denotes Expected Calibration Error, which is 
%a measure of calibration accuracy.
the weighted-average gap between model confidence and actual accuracy. (Gap is shown in green bars.)
Conventional attention models result in poorly calibrated networks while our UA yields a well-calibrated one. Such accurately calibrated networks allow us to perform reliable prediction by leveraging prediction confidence to decide whether to predict or defer prediction.}\label{fig:reliability_diagram}
\vspace{-0.2in}
\end{figure*}

Attention mechanism~\cite{Bahdanau15} is an effective means of guiding the model to focus on a partial set of most relevant features for each input instance. It works by generating (often sparse) coefficients for the given features in an input-adaptive manner, to allocate more weights to the features that are found to be relevant for the given input. Attention mechanism has been shown to significantly improve the model performance for machine translation~\cite{Bahdanau15} and image annotation~\cite{show_attend_tell} tasks. Another important feature of the attention mechanism is that it allows easy interpretation of the model via the generated attention allocations, and one recent work on healthcare domain~\cite{retain} is focusing on this aspect. 

Although interpretable, attention mechanisms are still limited as means of implementing safe deep learning models for safety-critical tasks, as they are not necessarily reliable. The attention strengths are commonly generated from a model that is trained in a weakly-supervised manner, and could be incorrectly allocated; thus they may not be safe to base final prediction on. To build a reliable model that can prevent itself from making critical mistakes, we need a model that knows its own limitation - \emph{when it is safe to make predictions and when it is not}. However, existing attention model cannot handle this issue as they do not have any notion of predictive uncertainty. This problem is less of an issue in the conventional use of attention mechanisms, such as machine translation or image annotation, where we can often find clear link between the attended parts and the generated output. However, when working with variables that are often noisy and may not be one-to-one matched with the prediction, such as in case of risk predictions with electronic health records, the overconfident and inaccurate attentions can lead to incorrect predictions (See Figure 1).

%, and often generates incorrect attentions that are interpreted in an unexpected way

To tackle this limitation of conventional attention mechanisms, we propose to allow the attention model to output uncertainty on each feature (or input) and further leverage them when making final predictions. Specifically, we model the attention weights as Gaussian distribution with \emph{input-dependent noise}, such that the model generates attentions with small variance when it is confident about the contribution of the given features, and allocates noisy attentions with large variance to uncertain features, for each input. This input-adaptive noise can model \emph{heteroscedastic uncertainty}~\cite{what_uncertainty} that varies based on the instance, which in turn results in uncertainty-based attenuation of attention strength.  We formulate this novel uncertainty-aware attention (UA) model under the Bayesian framework and solve it with variational inference.

We validate UA on tasks such as sepsis prediction in ICU and disease risk prediction from electronic health records (EHR) that have large degree of uncertainties in the input, on which our model outperforms the baseline attention models by large margins. Further quantitative and qualitative analysis of the learned attentions and their uncertainties show that our model can also provide richer interpretations that align well with the clinician's interpretations. For further validation on prediction reliability, we evaluate it for the uncertainty calibration performance, and prediction under the scenario where the model can defer the decision by saying ``I don't know'', whose results show that UA yields significantly better calibrated networks that can better avoid making incorrect predictions on instances that it is uncertain, compared to baseline attention models.

Our contribution in this paper is threefold:
\vspace{-0.1in}
\begin{itemize}
%\item We tackle the relatively less explored problem of achieving reliability in both the model interpretations and predictions, with deep neural networks. 
\item We propose a novel variational attention model with instance-dependent modeling of variance, that captures input-level uncertainty and use it to attenuate attention strengths. 
\item We show that our uncertainty-aware attention yields accurate calibration of model uncertainty as well as attentions that aligns well with human interpretations.
\item We validate our model on six real-world risk prediction problems in healthcare domains, for both the original binary classification task and classification with ``I don't know" decision,  and show that our model obtains significant improvements over existings attention models. 
\end{itemize}

\section{Related Work}

\paragraph{Prediction reliability}
There has been work on building a \emph{reliable} deep learning model\cite{timeseries_uncertainty_uber,bayesian_segnet,what_uncertainty}; that is, a deep network that can avoid making incorrect predictions when it is not sufficiently certain about its prediction. To achieve this goal, a model should know the limitation in the data, and in itself. One way to quantify such limitations is by measuring the \emph{predictive uncertainty} using Bayesian models. Recently, \cite{dropout_as_bayesian, rnn_dropout, cnn_dropout} showed that deep networks with dropout sampling \cite{dropout} can be understood as Bayesian neural networks. To obtain better calibrated dropout uncertainties, \cite{variational_dropout,concrete_dropout} proposed to automatically learn the dropout rates with proper reparameterization tricks \cite{concrete_distribution, vae}. While the aformentioned work mostly focus on accurate calibration of uncertainty itself, \cite{what_uncertainty} utilized dropout sampling to model predictive uncertainty in computer vision \cite{bayesian_segnet}, and also modeled label noise with learned variances, to implicitly attenuate loss for the highly uncertain instances. Our work has similar motivation, but we model the uncertainty in the \emph{input data} rather than in labels. By doing so, we can accurately calibrate deep networks for improved reliability. \cite{input_uncertainty_augmentation} has a similar motivation to ours, but with different applications and approaches. 
%There exists quite a few work on this topic recently~\cite{guo17,deep_ensembles}.
There exists quite a few work about uncertainty calibration and its quantification.
\cite{guo17} showed that the modern deep networks are poorly calibrated despite their accuracies, and proposed to tune factors such as depth, width, weight decay for better calibration of the model, and~\cite{deep_ensembles} proposed ensemble and adversarial training for the same objective. 

%which allows easy interpretation of the complex deep learning models via the learned attentions. Choi et al. mainly focus on this interpretability of the attention 

\paragraph{Attention mechanism}
%Attention mechanism~\cite{Bahdanau15} is an approach to adaptively select only the partial set of features (or inputs) for each input, such that the model input-adaptively focuses on more relevant features for prediction. 
The literature on the attention mechanism is vast, which includes its application to machine translation~\cite{Bahdanau15}, memory-augmented networks~\cite{e2ememnet}, and for image annotation~\cite{show_attend_tell}. Attention mechanisms are also used for interpretability, as in Choi et al.~\cite{retain} which proposed a RNN-based attention generator for EHR that can provide attention on both the hospital visits and variables for further analysis by clincians. Attentions can be either deterministic or probabilistic, and soft (non-sparse) or hard (sparse). Some probabilistic attention models~\cite{show_attend_tell} use variational inference as used in our model. However, while their direct learning of multinoulli distribution only considers whether to attend or not without consideration of variance, our attention mechanism models varying degree of uncertainty for each input by input-dependent learning of attention noise (variance).

%\paragraph{Variational inference for DNNs}
%Variational inference~\cite{peterson87} is a technique to solve intractable posterior estimation problem by fitting a parametric distribution to it. Finding its optimal parameter then can be done by maximizing the evidence lower bound (ELBO), but obtaining analytic solutions is difficult for complex models such as deep networks. To tackle this challenge, \cite{vae} recently proposed a reparametrization trick that enables to solve ELBO with stochastic gradient descent. \cite{cvae} proposed a conditional variational inference framework, which learns the parametric distribution $q$ conditioned on the input data and the label. Our model uses this conditional framework as the attention should be generated input-adaptively for each given data instance.

\paragraph{Risk analysis from electronic health records} Our work is mainly motivated by the needs of performing reliable risk prediction with electronic health records. There exists plentiful prior work on this topic, but to mention a few, Choi et al.~\cite{retain} proposed to predict heart failure risk with attention generating RNNs and Futoma et al.~\cite{futoma17} proposed to predict sepsis using RNN, preprocssing the input data using multivariate GP to resolve uneven spacing and missing entry problems. %This can be viewed as another way of input-level uncertainty modeling, while still orthogonal to our work.

\section{Approach}
We now describe our uncertainty-aware attention model. Let $\D$ be a dataset containing a set of $N$ input data points $\bX = [\bx^{(1)}\dots\bx^{(N)}]$ and the corresponding labels, $\bY = [\by^{(1)}\dots\by^{(N)}]$. For notational simplicity, we suppress the data index $n=1,\dots,N$ when it is clear from the context.
%For example, if the task is image annotation~\citep{show_attend_tell}, then $\{\bX, \by}$ takes

%Our goal is to learn stochastic attentions \emph{having low variances}. The existing deterministic attention mechanisms ignore this variance, and simply finds the best mapping that minimizes a given loss function. The idea is, if it is reliable, then the variance should be limited. We do not want attentions to have high variance if 

We first present a general framework of a stochastic attention mechanism. Let $\bv(\bx) \in \real^{r\times i}$ be the concatenation of $i$ intermediate features, each column of which $\bv_j(\bx)$ is a length $r$ vector, from an arbitrary neural network. From $\bv(\bx)$, a set of random variables $\{\ba_j\}_{j=1}^i$ is conditionally generated from some distribution $p(\ba|\bx)$ where the dimension of $\ba_j$ depends on the model architecture.  Then, the context vector $\bc \in \real^r$ is computed as follows:
% and element-wisely multiplied back to $\bv(\bx)$ along axis 1. Then summation along axis 1 results in a set of context vectors $\bc \in \real^r$, on which we perform final prediction $\hat{\by}$.
\begin{align*}
	\bc(\bx) = \sum_{j=1}^i \ba_j \odot \bv_j(\bx),\quad\hat{\by} = f(\bc(\bx)) 
\end{align*}
where the operator $\odot$ is properly defined according to the dimensionality of $\ba_j$; if $\ba_j$ is a scalar, it is simply the multiplication while for $\ba_j \in \real^r$, it is the element-wise product. The function $f$ here produces the prediction $\hat{\by}$ given the context vector $\bc$. 

The attention could be generated either deterministically, or stochastically. The stochastic attention mechanism is proposed in~\cite{show_attend_tell}, where they generate $\ba_j \in \{0,1\}$ from Bernoulli distribution. This variable is learned by maximizing the evidence lower bound (ELBO) with additional regularizations for reducing variance of gradients. In~\cite{show_attend_tell}, the stochastic attention is shown to perform better than the deterministic counterpart, on image annotation task.

\subsection{Stochastic attention with input-adaptive Gaussian noise}
Despite the performance improvement in \cite{show_attend_tell}, there are two limitations in modeling stochastic attention directly with Bernoulli (or Multinoulli) distribution as \cite{show_attend_tell} does, in our purposes:

\textbf{1) The variance $\sigma^2$ of Bernoulli is completely dependent on the allocation probability $\mu$.}

Since the variance for Bernoulli distribution is decided as $\sigma^2 = \mu(1-\mu)$, the model thus cannot generate $\ba$ with low variance if $\mu$ is around $0.5$, and vice versa. 
To overcome such limitation, we disentangle the attention strength $\ba$ from the attention uncertainty so that the uncertainty could vary even with the same attention strength. 

\textbf{2) The vanilla stochastic attention models the noise independently of the input.} 

This  makes it infeasible to model the amount of uncertainty for each input, which is a crucial factor for reliable machine learning. Even for the same prediction tasks and for the same set of features, the amount of uncertainty for each feature may largely vary across different instances. %For instance, \hb{to be filled out..}

To overcome these two limitations, we model the standard deviation $\sigma$, which is indicative of the uncertainty, as an \emph{input-adaptive} function $\sigma(\mbf{x})$, enabling to reflect different amount of confidence the model has for each feature, for a given instance. As for distribution, we use \emph{Gaussian} distribution, which is probably the most simple and efficient solution for our purpose, and also easy to implement.

We first assume that a subset of the neural network parameters $\bs\omega$, associated with generating attentions, has zero-mean isotropic Gaussian prior with precision $\tau$. Then the attention scores before squashing, denoted as $\bz$, are generated from conditional distribution $p_\theta(\bz|\bx,\bs\omega)$, which is also Gaussian:
%Specifically, we propose to use Gaussian additive noise instead of Bernoulli as the distribution of attention.
%Specifically, we use Gaussian distribution instead of Bernoulli distribution, in order to express different amount of variances for different input points, even with same mean:
\begin{align}
	p(\bs\omega) = \N(\mathbf{0}, \tau^{-1}\bI),\quad
	p_\theta(\bz|\bx,\bs\omega) = \N(\bs\mu(\bx,\bs\omega;\theta),\diag(\bs\sigma^2(\bx,\bs\omega;\theta))) \label{eq:uncertainty}
\end{align}
where $\bs\mu(\cdot,\bs\omega;\theta)$ and $\bs\sigma(\cdot,\bs\omega;\theta)$ are mean and s.d., parameterized by $\theta$. Note that 
%Gaussian is on the input to the attentions in Eq. \eqref{eq:aleatoric_uncertainty}, followed by  
$\bs\mu$ and $\bs\sigma$ are generated from the same layer, but with different set of parameters, although we denote those parameters as $\theta$ in general.
%In the above equation, we assumed that we use a common, shared set of $\theta$ for generating $\mu$ and $\sigma$. The motivation is that $\mu^p$ and $\sigma^p$ should be generated in relation to others, as if the variation is high ($\sigma$ is high) then this means that the given input is uncertain and $\mu$ should be lowered to reflect it. However they could be modeled with separate sets of parameters as well.
The actual attention $\ba$ is then obtained by applying some squashing function $\pi(\cdot)$ to $\bz$ (e.g. sigmoid or hyperbolic tangent):
%\begin{align}
	$\ba = \pi(\bz)$. %\label{eq:squashing}
%\end{align}
%Also note that $\mu$ and $\sigma$ are modeled with different set of parameters, although they are denote as general $\theta$.
%(Eq. \eqref{eq:epistemic_uncertainty}), whose role will be discussed in Section~\ref{section:two_uncertainties} in more detail. 
For comparison, one can think of the vanilla stochastic attention of which variance is independent of inputs.
\begin{align}
p(\bs\omega) = \N(\mathbf{0}, \tau^{-1}\bI),\quad
p_\theta(\bz|\bx,\bs\omega) = \N(\bs\mu(\bx,\bs\omega;\theta),\diag(\bs\sigma^2)) \label{eq:vanila_stochastic_attention}
\end{align}
However, as we mentioned, this model cannot express different amount of uncertainties over features.

One important aspect of our model is that, in terms of graphical representation, the distribution $p(\bs\omega)$ is independent of $\bx$, while the distribution $p_\theta(\bz|\bx,\bs\omega)$ is conditional on $\bx$. That is, $p(\bs\omega)$ tends to capture uncertainty of model parameters (epistemic uncertainty), while $p_\theta(\bz|\bx,\bs\omega)$ reacts sensitively to uncertainty in data, varying across different input points (heteroscedastic uncertainty)~\cite{what_uncertainty}. 
%These two uncertainties are not clearly separable as neither of them can be used alone as posterior process that captures both model and data uncertainties. However, 
When \emph{modeled together}, it has been empirically shown that the quality of uncertainty improves~\cite{what_uncertainty}. Such modeling both input-agnostic and input-dependent uncertainty is especially important in risk analysis tasks in healthcare, to capture both the uncertainty from insufficient amount of clinical data (e.g. rare diseases), and the uncertainty that varies from patients to patients (e.g. sepsis).
%epistemic uncertainty mainly captures uncertainty from lack of data, while heteroscedastic uncertainty captures intrinsic noise in data (e.g., noise in label) varying across different input points~\cite{what_uncertainty}. 

%\subsection{Uncertainty modeling}\label{section:two_uncertainties}
%\hb{input dependent uncertainty is important...}
\subsection{Variational inference}
We now model what we have discussed so far. Let $\bZ$ be the set of latent variables $\{\bz^{(n)}\}_{n=1}^{N}$ that stands for attention weight before squashing. In neural network, the posterior distribution $p(\bZ,\bs\omega|\D)$ is usually computationally intractable since $p(\D)$ is so due to nonlinear dependency between variables. Thus, we utilize variational inference, which is an approximation method that has been shown to be successful in many applications of neural networks~\cite{vae,cvae}, along with reprameterization tricks for pathwise backpropagation~\cite{variational_dropout,concrete_dropout}. 

Toward this, we first define our variational distribution as
\begin{equation}
\begin{aligned}
q(\bZ,\bs\omega|\D) = q_\bM(\bs\omega|\bX,\bY)q(\bZ|\bX,\bY,\bs\omega).
\end{aligned}
\end{equation}
We set $q_\bM(\bs\omega|\bX,\bY)$ to dropout approximation \cite{dropout_as_bayesian} with variational parameter $\bM$. \cite{dropout_as_bayesian} showed that a neural network with Gaussian prior on its weight matrices can be approximated with variational inference, in the form of dropout sampling of deterministic weight matrices and $\ell_2$ weight decay.
For the second term, we drop the dependency on $\bY$ (since it is not available in test time) and simply set $q(\bZ|\bX,\bY,\bs\omega)$ to be equivalent to $p_\theta(\bZ|\bX,\bs\omega)$, which works well in practice~\citep{cvae,show_attend_tell}.
%as decomposable by each instance and each element, which follows univariate gaussian with another set of parameters $\phi$.
%\begin{equation}
%\begin{aligned}
%q_\phi(\bZ|\bX,\bY,\bs\omega) &= \prod_{n=1}^{N} \prod_{j=1}^{i} q_\phi\left(z_j^{(n)}|\bx^{(n)},\by^{(n)},\bs\omega\right) \\
%q_\phi(z_j|\bx,\by,\bs\omega) &= \N\left(\mu_j(\bx,\by,\bs\omega;\phi), \sigma_j(\bx,\by,\bs\omega;\phi)^2\right)
%\label{eq:q_gaussian}
%\end{aligned}
%\end{equation}

Under the SGVB framework \cite{vae}, we 
%then minimize $\kl[q_\phi(\bZ,\bs\omega|\D)\|p(\bZ,\bs\omega|\D)]$. It is known to be equivalent to 
maximize the evidence lower bound (ELBO):
\begin{align}
\log p(\bY|\bX) \geq \ &\E_{\bs\omega\sim q_\bM(\bs\omega|\bX,\bY), \bZ \sim p_\theta(\bZ|\bX,\bs\omega)}\left[\log p(\bY|\bX,\bZ,\bs\omega)\right]
\label{eq:expected_log_likelihood} \\
&- \kl[q_\bM(\bs\omega|\bX,\bY)\|p(\bs\omega)] - \kl[q(\bZ|\bX,\bY,\bs\omega)\|p_\theta(\bZ|\bX,\bs\omega)]
\label{eq:kl}
\end{align}
%where $\theta$ is parameters associated with generating $\bZ$ from the prior distribution $p(\bZ|\bX,\bs\omega)$. 
where we approximate the expectation in \eqref{eq:expected_log_likelihood} via Monte-Carlo sampling. The first KL term nicely reduces to $\ell_2$ regularization for $\bM$ with dropout approximation~\cite{dropout_as_bayesian}. 
The second KL term vanishes as the two distributions are equivalent.
%KL term in Eq.~\eqref{eq:kl_aleatoric} can be decomposed as sum of individual instance and elements, which is simply the KL divergence between two univariate Gaussians (Eq. \ref{eq:aleatoric_uncertainty}, Eq.\ref{eq:q_gaussian}):
%\begin{equation}
%\begin{aligned}
%\kl[q_\phi\|p_\theta] = \log\frac{\sigma^q}{\sigma^p} + \frac{(\sigma^p)^2 + (\mu^p-{\mu^q})^2}{2(\sigma^q)^2} + const.\label{eq:kl_normal}
%\end{aligned}
%\end{equation}
%where we denote $\{\mu_\phi, \sigma_\phi\}$ from $q_\phi$ and $\{\mu_\theta, \sigma_\theta\}$ from $p_\theta$, for notational simplicity.
Consequently, our final maximization objective is:
%\begin{equation}
%\begin{aligned}
%\loss&(\theta,\phi,\bM;\bX,\bY) =\sum_{n=1}^{N}\Big\{\log p_\theta(\by^{(n)}|\tilde{\bz}^{(n)},\bx^{(n)}) \\
% &- \kl\left[q_\phi(\bz^{(n)}|\bx^{(n)},\by^{(n)})\|p_\theta(\bz^{(n)}|\bx^{(n)})\right] \Big\} - \lambda \|\bM\|^2
%\label{eq:each_obj}
%\end{aligned}
%\end{equation}
\begin{align}
\loss(\theta,\bM;\bX,\bY) =\sum \log p_\theta(\by^{(n)}|\tilde{\bz}^{(n)},\bx^{(n)}) - \lambda \|\bM\|^2 \label{eq:each_obj}
\end{align}
where we first sample random weights with dropout masks
$\widetilde{\bs\omega} \sim q_\bM(\bs\omega|\bX,\bY)$ and sample $\bz$ such that $\tilde{\bz} = g(\bx,\tilde{\bs\varepsilon},\widetilde{\bs\omega}), \tilde{\bs\varepsilon} \sim \N(\mathbf{0},\bI)$, with a pathwise derivative function $g$ for reparameterization trick. $\lambda$ is a tunable hyperparameter; however in practice it can be simply set to common $\ell_2$ decay shared throughout the network, including other deterministic weights.

When testing with a novel input instance $\bx^*$, we can compute the probability of having the correct label $y^*$ by our model, $p(\by^*|\bx^*)$ with Monte-Carlo sampling: 
\begin{align}
p(\by^*|\bx^*) = \iint p(\by^*|\bx^*,\bz)p(\bz|\bx^*,\bs\omega)p(\bs\omega|\bX,\bY) \text{d}\bs\omega \text{d}\bz \approx \frac{1}{S}\sum_{s=1}^{S} p(\by^*|\bx^*,\tilde{\bz}^{(s)}) \label{eq:test}
\end{align}
where we first sample dropout masks $\widetilde{\bs\omega}^{(s)} \sim\ q_\bM(\bs\omega|\bX,\bY)$ and then sample $\tilde\bz^{(s)} \sim p_\theta(\bz|\bx^*,\widetilde{\bs\omega}^{(s)})$.
%Alternatively, we can simply put the expectation inside $p(\by^*|\bx^*) \approx p(\by^*|\bx^*,\E[\bz|\bx^*,\E[\bs\omega]])$ which is a usual practice for many practitioners and works well in practice.

%In mini-batch framework, we randomly sample $B$ training instances and corresponding $\tilde{\bs\varepsilon}$ every iteration, compute \eqref{eq:each_obj} for all those samples, and optimize via stochastic gradient descent. 
\paragraph{Uncertainty Calibration} The quality of uncertainty from \eqref{eq:test} can be evaluated with reliability diagram shown in Figure \ref{fig:reliability_diagram}. Better calibrated uncertainties produce smaller \emph{gaps} beween model confidences and actual accuracies, shown in green bars. Thus, the perfect calibration occurs when the confidences exactly matches the actual accuracies: $p(\text{correct}|\text{confidence}=\rho)=\rho, \forall\rho\in[0,1]$ \cite{guo17}. Also, \cite{ece,guo17} proposed a summary statistic for calibration, called the Expected Calibration Error (ECE). It is the expected \emph{gap} w.r.t. the distribution of model confidence (or frequency of bins):
\begin{align}
	\text{ECE} = \E_\text{confidence}\big[|p(\text{correct}|\text{confidence})-\text{confidence}|\big]\label{eq:ece}
\end{align}

\section{Application to classification from time-series data}
Our variational attention model is generic and can be applied to any generic deep neural network that leverages attention mechanism. However, in this section, we describe its application to prediction from time-series data, since our target application is risk analysis from electronic health records.

\paragraph{Review of the RETAIN model}
As a base deep network for learning from time-series data, we consider RETAIN~\cite{retain}, which is an attentional RNN model with two types of attentions--across \emph{timesteps} and across \emph{features}. RETAIN obtains state-of-the-art performance on risk prediction tasks from electronic health records, and is able to provide useful interpretations via learned attentions. 
%We first introduce the notation for \cite{retain}. Suppose there are patients $n=1,...,N$. Each patient $n$ has timestep $i=1,...,T(n)$, and the corresponding set of input features up to timestep $i$ is denoted as $\bbX_{i}^{(n)}=\left\{\bx_1^{(n)},...,\bx_i^{(n)}\right\}$, where $\bx \in \real^{r}$. The corresponding set of binary labels is only provided at the last timestep $T{(n)}$, denoted as $y^{(n)}$. The whole training dataset is then $\D = \left\{\left(\bx_1^{(n)},...,\bx_{T(n)}^{(n)}, y^{(n)}\right)\right\}_{n=1}^N$. When we are interested in timestep $i$, its intermediate timesteps are indexed as $j=1,...,i$. We will not specify patient index for variables if context is clear. 

We now briefly review the overall structure of RETAIN. We match the notation with those in the original paper for clear reference. Suppose we are interested in a timestep $i$. With the input embeddings $\bv_1,\dots,\bv_i$, we generate two different attentions: across timesteps ($\alpha$) and features ($\bbeta$).
\begin{align}
&\bg_i,...,\bg_1 = \rnn_\alpha (\bv_i,...,\bv_1;\bs\omega), 
&\bh_i,...,\bh_1 = \rnn_\bbeta (\bv_i,...,\bv_1;\bs\omega),\label{eq:rnn} \\
&e_j = \bw_\alpha^\T \bg_j + b_\alpha\ \ \text{for}\ \ j=1,...,i, 
&\bd_j = \bW_\bbeta \bh_j + \bb_\bbeta\ \ \text{for}\ \ j=1,...,i,\label{eq:retain_logit} \\
&\alpha_1,...,\alpha_i = \softmax(e_1,...,e_i), 
&\bbeta_j = \tanh(\bd_j)\ \ \text{for}\ \  j=1,...,i.\label{eq:retain}
\end{align}
%\begin{align}
%\alpha_1,\alpha_2,...,\alpha_i = \softmax(e_1,e_2,...,e_i), \quad
%\bbeta_j = \tanh(\bd_j)\ \ \text{for}\ \ j=1,...,i.\label{eq:retain}
%\end{align}
The parameters of two RNNs are collected as $\bs\omega$. From the RNN outputs $\bg$ and $\bh$, the attention logits $e$ and $\bd$ are generated, followed by squashing functions $\softmax$ and $\tanh$ respectively. %$\bd_1,\dots,\bd_i$ denote logits of each attention repectively, generated by two different RNNs with the same input $\bv_1,\dots,\bv_i$.
%$\rnn(\cdot)$ denotes a generic recurrent neural networks, such as LSTM and GRU.
%
%Another attention $\bbeta_j \in [-1,1]^r$ for timestep $j$ and across features $k=1,...,r$ is generated as
%\begin{equation}
%\begin{aligned}
%\bh_i,\bh_{i-1},...,\bh_1 &= \rnn_\bbeta (\bv_i,\bv_{i-1},...,\bv_1;\bs\omega) \\
%\bd_j &= \bW_\bbeta \bh_j + \bb_\bbeta \\
%\bbeta_j &= \tanh(\bd_j)\ \ \ \text{for}\ \ \ j=1,...,i.
%\label{eq:beta}
%\end{aligned}
%\end{equation}
Then the generated two attentions $\alpha$ and $\bbeta$ are multiplied back to the input embedding $\bv$, followed by a convex sum $\bc$ up to timestep $i$: $\bc_i = \sum_{j=1}^{i}\alpha_j\bbeta_j\odot\bv_j$. A final linear predictor is learned based on it: $\widehat{y}_i = \sigm(\bw^\T\bc_i + b)$.

%\begin{align}
%\bc_i = \sum_{j=1}^{i}\alpha_j\bbeta_j\odot\bv_j,\quad
%\widehat{y}_i = \sigm(\bw^\T\bc_i + b)
%\label{eq:final_prediction}
%\end{align}

%\begin{equation}
%\begin{aligned}
%\frac{1}{N}\sum_{n=1}^{N}
%\left\{
%y^{(n)}\log(\widehat{y}^{(n)})
%+(1-y^{(n)})\log(1-\widehat{y}^{(n)})
%\right\}\nonumber
%\end{aligned}
%\end{equation}
%Note that unlike in~\citep{retain}, the label is only provided at the end of the timestep for each patient.

The most important feature of RETAIN is that it allows us to interpret what the model has learned as follows. What we are interested in is \emph{contribution}, which shows $x_{k}$'s aggregate effect to the final prediction at time $j$. Since RETAIN has attentions on both timesteps ($\alpha_j$) and features ($\bbeta_{j}$), the computation of aggregate contribution takes both of them into consideration when computing the final contribution of an input data point at a specific timestep: $\omega(y,x_{j,k}) = \alpha_j \bw^\T (\bbeta_j\odot \bW_{emb}[:,k]) x_{j,k}$. 
%\begin{equation}
%\begin{aligned}
%\omega(y,x_{j,k}) = \alpha_j \bw^\T (\bbeta_j\odot \bW_{emb}[:,k]) x_{j,k}
%\label{eq:contribution}
%\end{aligned}
%\end{equation}
In other words, it is a certain portion of logit $\sigm^{-1}(\widehat{y}_i) = \bw^\T \bc_i + b$ for which $x_{j,k}$ is responsible.

\paragraph{Interpretation as a probabilistic model}
%For a given timestep $i$, we define the set of inputs for it as $\bX_i = [\bx_1...\bx_i]$. Suppose each patient $n$ has a sequence of record that consists of timesteps $t=1,...,T(n)$. Then,
%\begin{align}
%\D=\{\bX,\bY\},\quad
%\bX = \left\{\bX_{T(n)}^{(n)}\right\}_{n=1}^N,\quad
%\bY = \left\{y^{(n)}\right\}_{n=1}^N
%\nonumber
%\end{align}
The interpretation of RETAIN as a probabilistic model is quite straightforwrad. First, the RNN parameters $\bs\omega$ \eqref{eq:rnn} as gaussian latent variables \eqref{eq:uncertainty} are approximated with MC dropout with fixed probabilities \cite{dropout_as_bayesian, rnn_dropout, bayesian_lstm}. The input dependent latent variables $\bZ$ \eqref{eq:uncertainty} simply correspond to the collection of $e$ and $\bd$ \eqref{eq:retain_logit}, the attention logits. The log variances of $e$ and $\bd$ are generated in the same way as their mean, from the output of RNNs $\bg$ and $\bd$ but with different set of parameters. Also the reparameterization trick for diagonal gaussian is simple \cite{vae}.
We now maximize the ELBO \eqref{eq:each_obj}, equipped with all the components $\bX$,$\bY$,$\bZ$, and $\bs\omega$ as in the previous section.
%, and epistemic/heteroscedastic uncertainties can be sampled from Eq. \eqref{eq:sample_epistemic} and Eq. \eqref{eq:sample_aleatoric}.

\section{Experiments}
\vspace{-0.1in}
We validate the performance of our model on various risk prediction tasks from multiple EHR datasets, for both the prediction accuracy (Section 5.3) and prediction reliability (Section 5.4).

\subsection{Tasks and datasets}
%We test our model on three different EHR datasets with six different binary classification tasks. 

%each of which contains $80\%$ of the samples for training, $10\%$ for test, and the remaining $10\%$ for validation. 

\paragraph{1) PhysioNet} This dataset~\cite{physio2012} contains 4,000 medical records from ICU\footnote{We only use the TrainingSetA, for which the labels were available}. Each record contains 48 hours of records, with 155 timesteps, each of which contains 36 physiolocial signals including \emph{heart rate}, \emph{repiration rate} and \emph{temperature}. The challenge comes with four binary classification tasks, namely, 1) \textit{Mortality prediction},  2) \textit{Length-of-stay less than 3 days:} whether the patient will stay in ICU for less than three days, 3) \textit{Cardiac conditon:} whether the patient will have a cardiac condition, and 4) \textit{Recovery from surgery:} whether the patient was recovering from surgery.

\paragraph{2) Pancreatic Cancer}This dataset is a subset of an EHR database consisting of anonymized medical check-up records from 2002 to 2013, which includes around 1.5 million records. We extract $3,699$ patient records from this database, among which $1,233$ are patients diagnosed of pancreatic cancer. The task here is to predict the onsets of pancreatic cancer in 2013 using the records from 2002 to 2012 ($11$ timesteps), that consists of 34 variables regarding general information (e.g., sex, height, past medical history, family history) as well as vital information (e.g., systolic pressure, hemoglobin level, creatinine level) and risk inducing behaviors (e.g., tobacco and alcohol consumption).

\paragraph{3) MIMIC-Sepsis} This is the subset of the MIMIC III dataset~\cite{mimic3_ref} for sepsis prediction, which consists of 58,000 hospital admissions for 38,646 adults over 12 years. We use a subset that consists of 22,395 records of patients over age 15 and stayed in ICUs between 2001 and 2012, among which 2,624 patients are diagnosed of sepsis. We use the data from the first 48 hours after admission (24 timesteps). For features at each timestep, we select 14 sepsis-related variables including arterial blood pressure, heart rate, FiO2, and Glass Coma Score (GCS), following the clinicians' guidelines. We use Sepsis-related Organ Failure Assessment scores (SOFA) to determine the onset of sepsis.

For all datasets, we generates five random splits of training/validation/test with the ratio of $80\%:10\%:10\%$. For more detailed description of the datasets, please see \textbf{supplementary file}. 

%\paragraph{4) Pediatric Intensive Care Unit (P-ICU)} This dataset consists of electronic health records from a pediatric ICU at a major hospital, which contains records of 1632 patients. The patients' age ranges from 2 months to 19 years, with the average age of 5.8 years. We extract 9 physiological markers including \emph{pulse}, \emph{temperature}, and \emph{systolic blood pressure} from ICU records for every hour. We consider the mortality prediction task on this dataset, using records that cover the first 12 hours after ICU admission, where each hour is considered a timestep. The number of positive cases here is $116$. 
%\input{maintable}

\subsection{Baselines}
\vspace{-0.1in}
We now describe our uncertainty-calibrated attention models and relevant baselines.

\textbf{1) RETAIN-DA:} The recurrent attention model in~\cite{retain}, which uses deterministic soft attention.\\
\textbf{2) RETAIN-SA:} RETAIN model with the stochastic hard attention proposed by~\cite{show_attend_tell}, that models the attention weights with multinoulli distribution, which is learned by variational inference.\\
\textbf{3) UA-independent:} The input-independent version of our uncertainty-aware attention model in \eqref{eq:vanila_stochastic_attention} whose variance is modeled indepently of the input.\\
\textbf{4) UA:} Our input-dependent uncertainty-aware attention model in \eqref{eq:uncertainty}.\\ 
\textbf{5) UA+:} The same as UA, but with additional modeling of input-adaptive noise at the final prediction as done in~\cite{what_uncertainty}, to account for output uncertainty as well.\\

For network configuration and hyperparameters, see \textbf{supplementary file}. We will also release the codes for reproduction of the results.
%\paragraph{Network configuration and parameters} We trained all the models using Adam~\cite{adam} optimizer with dropout regularization. We set the maximum iteration for Adam optimizer as $100,000$, and for other hyperparameters, we searched for the optimal values by cross-validation, within predefined ranges as follows: Mini batch size: $\{32, 64, 128, 256\}$, learning rate: $\{0.01, 0.001, 0.0001\}$, \textit{L-2} regularization: $\{0.02, 0.002, 0.0002, 0.0004\}$, and dropout rate $\{0.1, 0.15, 0.2, 0.25, 0.3, 0.4, 0.5\}$. See supplementary file for more details on the network configurations.

%\input{other_table}

\subsection{Evaluation of the binary classification performance}
\begin{table*}[t]
\begin{center}
\resizebox{\textwidth}{!}{
\begin{tabular}{| c | c c c c | c | c |}
\hline
& \multicolumn{4}{c|}{PhysioNet} & Pancreatic & MIMIC \\
\cline{2-5} 
& Mortality & Stay $< 3$ & Cardiac & Recovery & Cancer & Sepsis \\
\hline
%RNN & & & \\
RETAIN-DA~\cite{retain}                    & 0.7652$\pm$ 0.02     & 0.8515$\pm$ 0.02     & 0.9485$\pm$ 0.01     & 0.8830$\pm$ 0.01     & 0.8528$\pm$ 0.01     & 0.7965$\pm$ 0.01\\
RETAIN-SA~\cite{show_attend_tell}       & 0.7635$\pm$ 0.02     & 0.8412$\pm$ 0.02     & 0.9360$\pm$ 0.01     & 0.8582$\pm$ 0.02     & 0.8444$\pm$ 0.01     & 0.7695$\pm$ 0.02 \\
\hline
UA-Independent & 0.7764$\pm$ 0.01      & 0.8572$\pm$ 0.02     & 0.9516$\pm$ 0.01     & 0.8895$\pm$ 0.01     & 0.8533$\pm$ 0.03     & 0.8019$\pm$ 0.01\\
UA              & \bf 0.7827$\pm$ 0.02  & \bf 0.8628$\pm$ 0.02 & 0.9563$\pm$ 0.01     & 0.9049$\pm$ 0.01     & 0.8604$\pm$ 0.01     & 0.8017$\pm$ 0.01\\
UA+            & 0.7770$\pm$ 0.02      & 0.8577$\pm$ 0.01     & \bf 0.9612$\pm$ 0.01 & \bf 0.9074$\pm$ 0.01 & \bf 0.8638$\pm$0.02  & \bf 0.8114$\pm$ 0.01\\
\hline
%UCA-PseudoLabel & & & &\\
%UCA-TrueLabel & & & &\\
\end{tabular}
}
\caption{\small The multi-class classification performance on the three electronic health records datasets. The reported numbers are mean AUROC and standard errors for 95\% confidence interval over five random splits.}
\label{tbl:physionet}
\end{center}
\vspace{-0.15in}
\end{table*}

We first examine the prediction accuracy of baselines and our models in a standard setting where the model always makes a decision. Table~\ref{tbl:physionet} contains the accuracy of baselines and our models measured in area under the ROC curve (AUROC). We observe that UA variants significantly outperforms both RETAIN variants with either deterministic or stochastic attention mechanisms on all datasets. Note that RETAIN-SA, that generates attention from Bernoulli distribution, performs the worst. This may be because the model is primarily concerned with whether to attend or not to each feature, which makes sense when most features are irrelevant, such as with machine translation, but not in the case of clinical prediction where most of the variables are important. UA-independent performs significantly worse than UA or UA+, which demonstrates the importance of input-dependent modeling of the variance. Additional modeling of output uncertainty with UA+ yields performance gain in most cases.

\begin{figure*}[t!]
\begin{center}
\small
\vspace{-0.1in}
\resizebox{12cm}{!}{
\begin{tabular}{l | c c c c c c c c c}
& MechVent & DiasABP & HR & Temp & SysABP & FiO2 & MAP & Urine & GCS\\
\hline
35m 5s& 0 & 81 & 61 & 36.2 & 135 & 1 & 71 & N/A & 15 \\
38m10s&0 & 75 & 64 & 36.7 & 94 & 1 & 74  & N/A & 15 \\
\bf38m \bf55s \textbf{(current)} &\bf1 & \bf67 & \bf57 & \bf35.2 & \bf105 & \bf1 & \bf80 & \bf35 & \bf10 \\
\end{tabular}
}
\begin{tabular}{c c c}
\hspace{-0.1in}
\includegraphics[width=4cm]{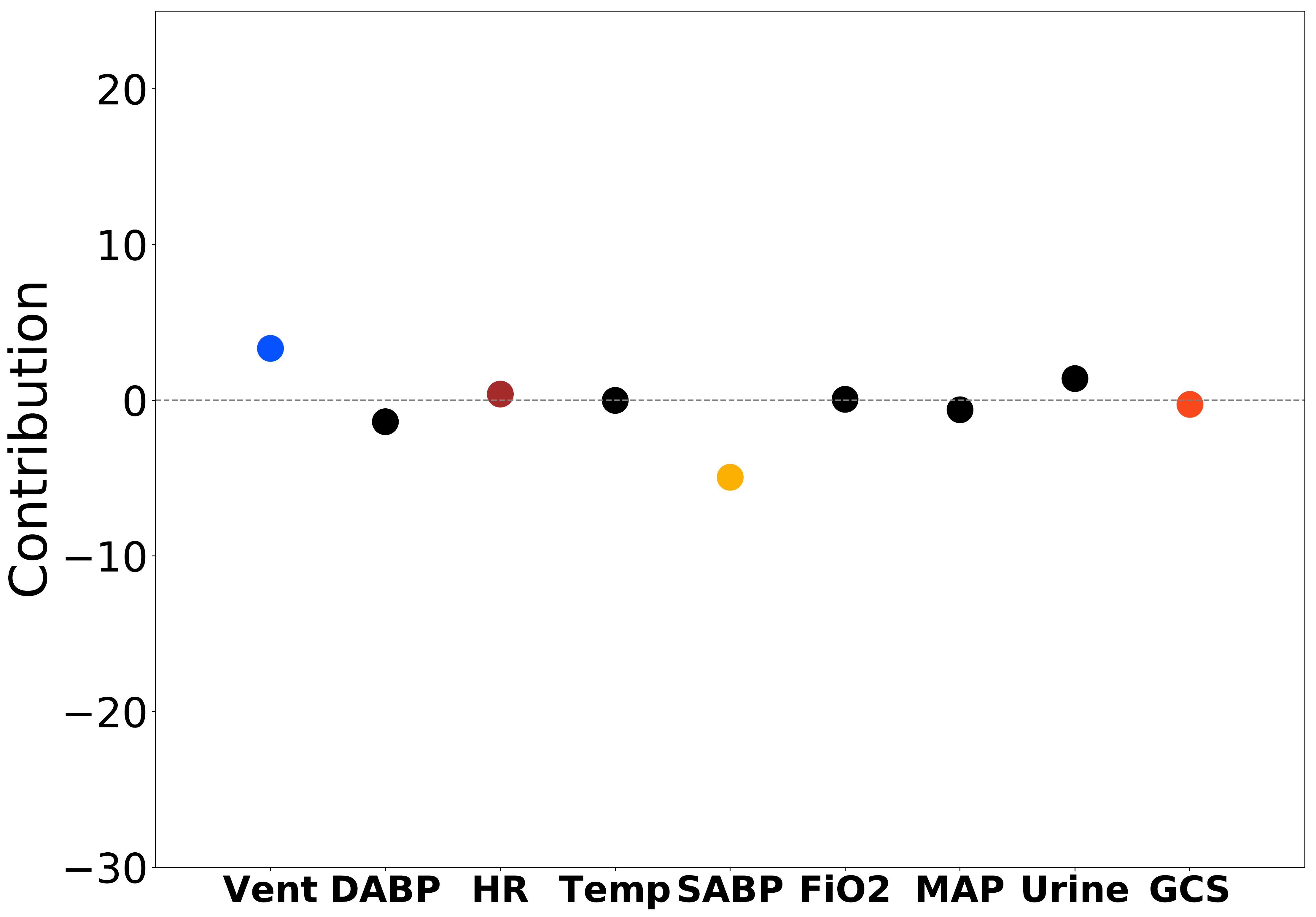}&
\hspace{-0.1in}
\includegraphics[width=4cm]{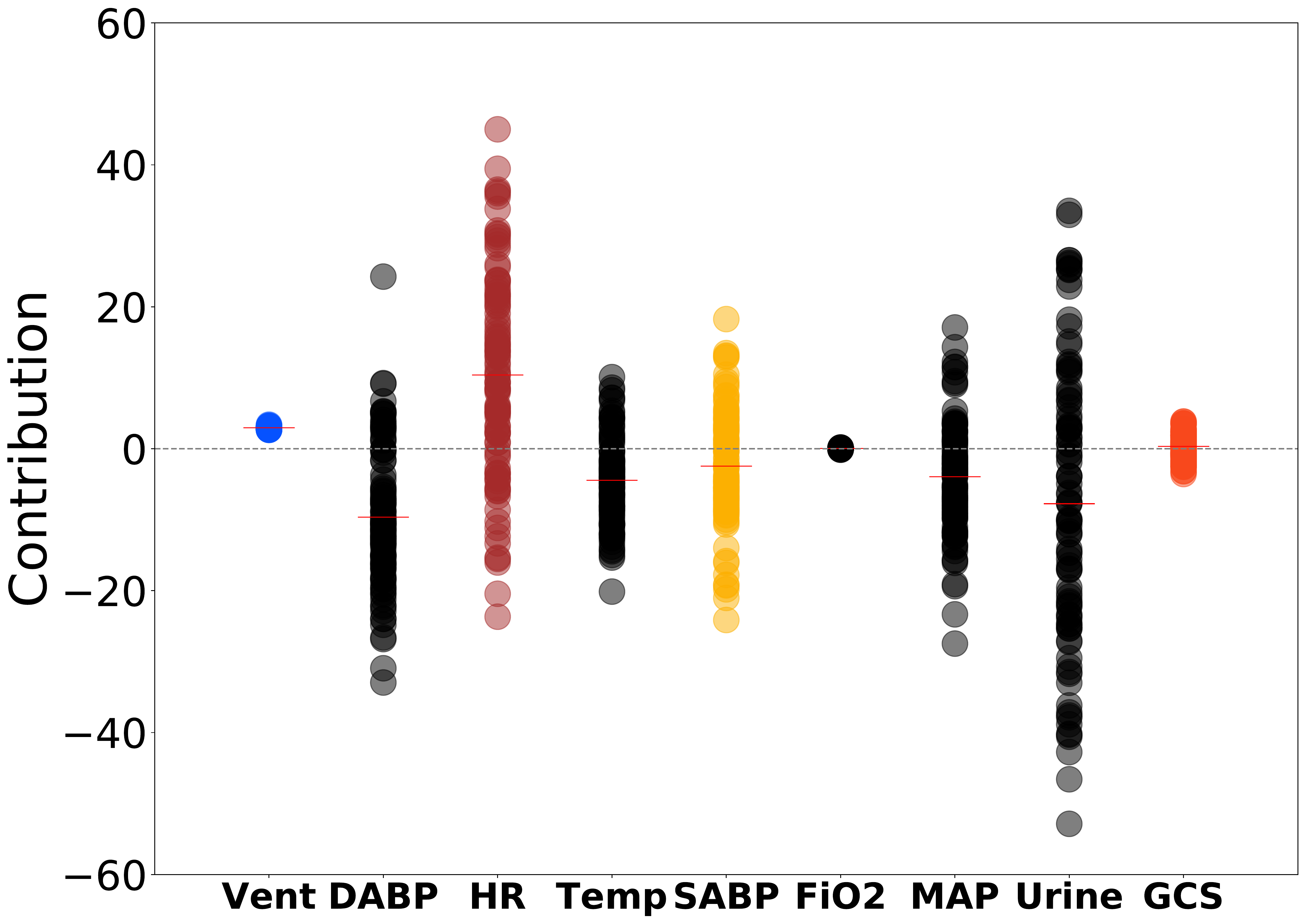}&
\hspace{-0.1in}
\includegraphics[width=4cm]{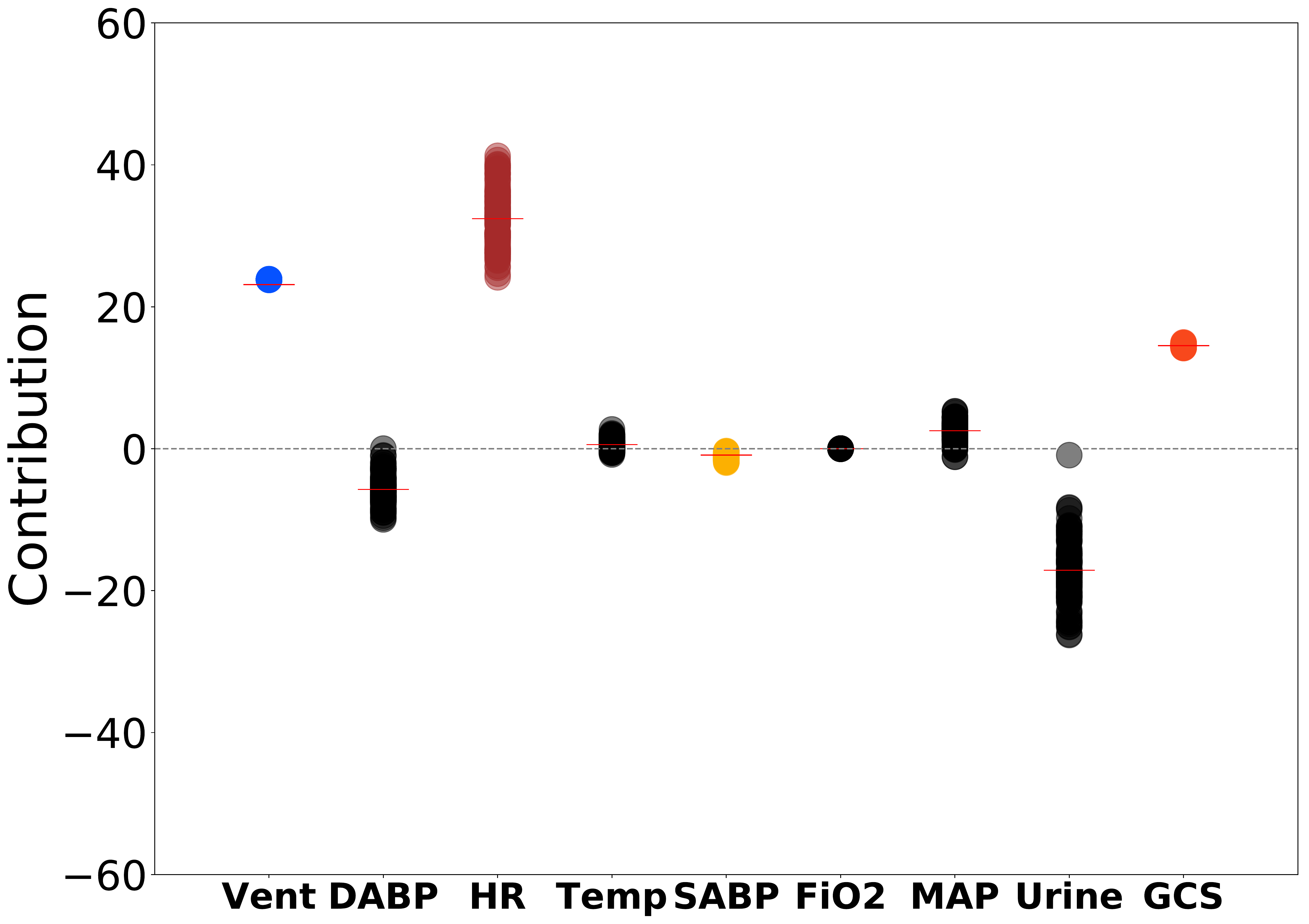}\\
(a) RETAIN & (b) RETAIN-SA & (c) UA \\
\end{tabular}
\end{center}
\vspace{-0.1in}
\caption{\small Visualization of contributions for a selected patient on PhysioNet mortality prediction task. \eat{Contribution indicates the extent to which each individual feature affects patient mortality.} \textbf{MechVent} - Mechanical ventilation, \textbf{DiasABP} - Diastolic arterial blood pressure, \textbf{HR} - Heart rate, \textbf{Temp} - Temperature, \textbf{SysABP} - Systolic arterial blood pressure, \textbf{FiO2} - Fractional inspired Oxygen, \textbf{MAP} - Mean arterial blood pressure, \textbf{Urine} - Urine output, \textbf{GCS} - Glasgow coma score. The table presents the value of physiological variables at the previous and the current timestep. Dots correspond to sampled attention weights.}\label{fig:attention}
\vspace{-0.1in}
\end{figure*}

\vspace{-0.2in}
\paragraph{Interpretability and accuracy of generated attentions} To obtain more insight, we further analyze the contribution of each feature in PhysioNet mortality task in Figure~\ref{fig:attention} for a patient at the timestep with the highest attention $\alpha$, with the help of a physician. The table in Figure~\ref{fig:attention} is the value of the variables at the previous checkpoints and the current timestep. 

The difference between the current and the previous tmesteps is significant - the patient is applied mechanical ventilation; the body temperature, diastolic arterial blood pressure, and heart rate dropped, and GCS, which is a measure of consciousness, dropped from 15 to 10. The fact that the patient is applied mechanical ventilation, and that the GCS score is lowered, are both very important markers for assessing patient's condition. Our model correctly attends to those two variables, with very low uncertainty. SysABP and DiasABP are variables that has cyclic change in value, and are all within normal range; however RETAIN-DA attended to these variables, perhaps due to having a deterministic model which led it to overfit. Heart rate is out of normal range (60-90), which is problematic but is not definitive, and thus UA attended to it with high variance. RETAIN-SA results in overly incorrect and noisy attention except for FiO2 that did not change its value. Attention on Urine by all models may be the artifact that comes from missing entry in the previous timestep. In this case, UA assigned high variance, which shows that it is uncertain about this prediction. %Please see supplementary file for more examples.

The previous example shows another advantage of our model: it provides a richer interpretations of why the model has made such predictions, compared to ones provided by deterministic or stochastic model without input-dependent modeling of uncertainty. This additional information can be taken account by clinicians when making diagnosis, and thus can help with prediction reliability. %However, it is not feasible for the clinicians to examine all the information given by the model. Thus, it would be helpful if the model can perform some screening and present only the most uncertain cases to them.  In next paragraph, we show how to utilize our model for this purpose.

\begin{wraptable}{r}{4.7cm}
\vspace{-0.15in}
%\begin{center}
\resizebox{4.5cm}{!}{

\begin{tabular}{| c | c | c |}

\hline
%RNN & & & \\
                     & \small Sensitivity  & \small Specificity \\
\hline
\small DA & \small 75$\%$       & \small 68$\%$   \\
\hline
\small UA                   & \small \bf 87$\%$   & \small \bf 82$\%$\\
\hline
\end{tabular}
}
\caption{\small Percentage of features selected from each model that match the features selected by the clinicians.}
\label{tbl:quantify}
%\end{center}
\vspace{-0.2in}
\end{wraptable}

We further compared UA against RETAIN-DA for accuracy of the attentions, using variables selected meaningful by clinicians as ground truth labels (avg. $132$ variables per record), from EHRs for a male and a female patient randomly selected from 10 age groups (40s-80s), on PhysioNet-Mortality. We observe that UA generates accurate interpretations that better comply with clinicians' intepretations (Table~\ref{tbl:quantify}).

\begin{table*}[t]
\begin{center}
\small
\resizebox{12cm}{!}{
\begin{tabular}{| c | c c c c | c | c |}
\hline
& \multicolumn{4}{c|}{PhysioNet} & Pancreatic & MIMIC \\
\cline{2-5} 
& Mortality & Stay $< 3$ & Cardiac & Recovery & Cancer & Sepsis \\
\hline
%RNN & & & \\
RETAIN-DA~\cite{retain}           & 7.23 $\pm$ 0.56      & 2.04 $\pm$ 0.56      & 5.70 $\pm$ 1.56      & 4.89 $\pm$ 0.97     & 5.45 $\pm$ 0.79     & 3.05 $\pm$ 0.56     \\
RETAIN-SA~\cite{show_attend_tell} & 7.70 $\pm$ 0.60      & 3.77 $\pm$ 0.07      & 8.82 $\pm$ 0.64      & 5.39 $\pm$ 0.80     & 9.69 $\pm$ 3.90     & 5.75 $\pm$ 0.29     \\
\hline
UA-Independent                    & 5.03 $\pm$ 0.94      & 2.74 $\pm$ 1.44      & 3.55 $\pm$ 0.56      & 4.87 $\pm$ 1.46     & 4.51 $\pm$ 0.72     & 2.04 $\pm$ 0.62     \\
UA                                & \bf 4.22 $\pm$ 0.82  & \bf 1.43 $\pm$ 0.53  & 3.33 $\pm$ 0.96      & 4.46 $\pm$ 0.73     & 3.61 $\pm$ 0.55     & \bf 1.78 $\pm$ 0.41 \\
UA+                               & 4.41 $\pm$ 0.52      & 1.68 $\pm$ 0.16      & \bf 2.66 $\pm$ 0.16  & \bf 3.98 $\pm$ 0.59 & \bf 3.22 $\pm$ 0.69 & 2.04 $\pm$ 0.62     \\
\hline
%UCA-PseudoLabel & & & &\\
%UCA-TrueLabel & & & &\\
\end{tabular}
}
\\
\caption{\small Mean Expected Calibration Error (ECE) of various attention models over 5 random splits.}
\label{tbl:ece}
\end{center}
\vspace{-0.15in}
\end{table*}

\begin{figure*}
\small
\begin{tabular}{c c c c c c}
\hspace{-0.14in}
\includegraphics[height=2.2cm]{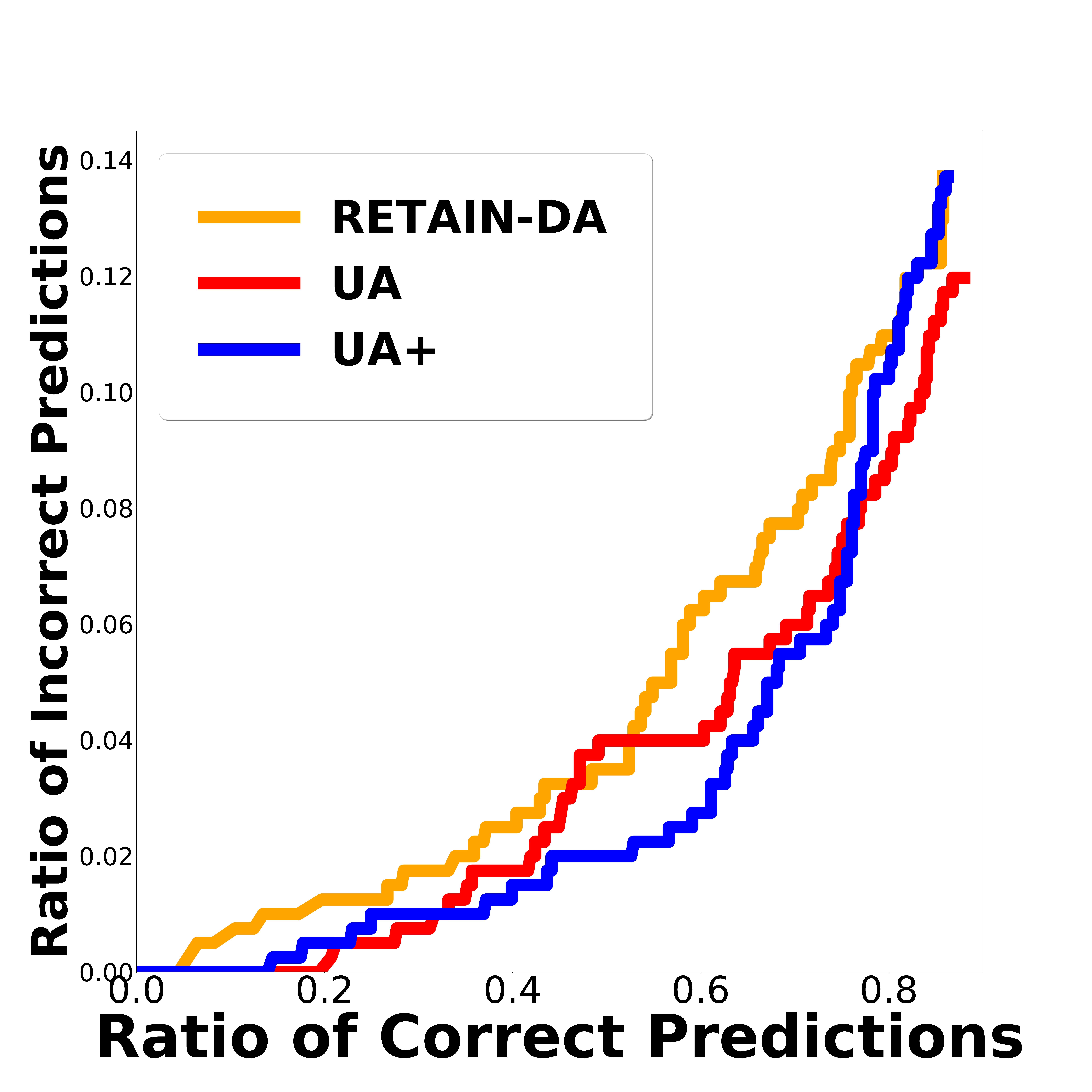}&
\hspace{-0.12in}
\includegraphics[height=2.2cm]{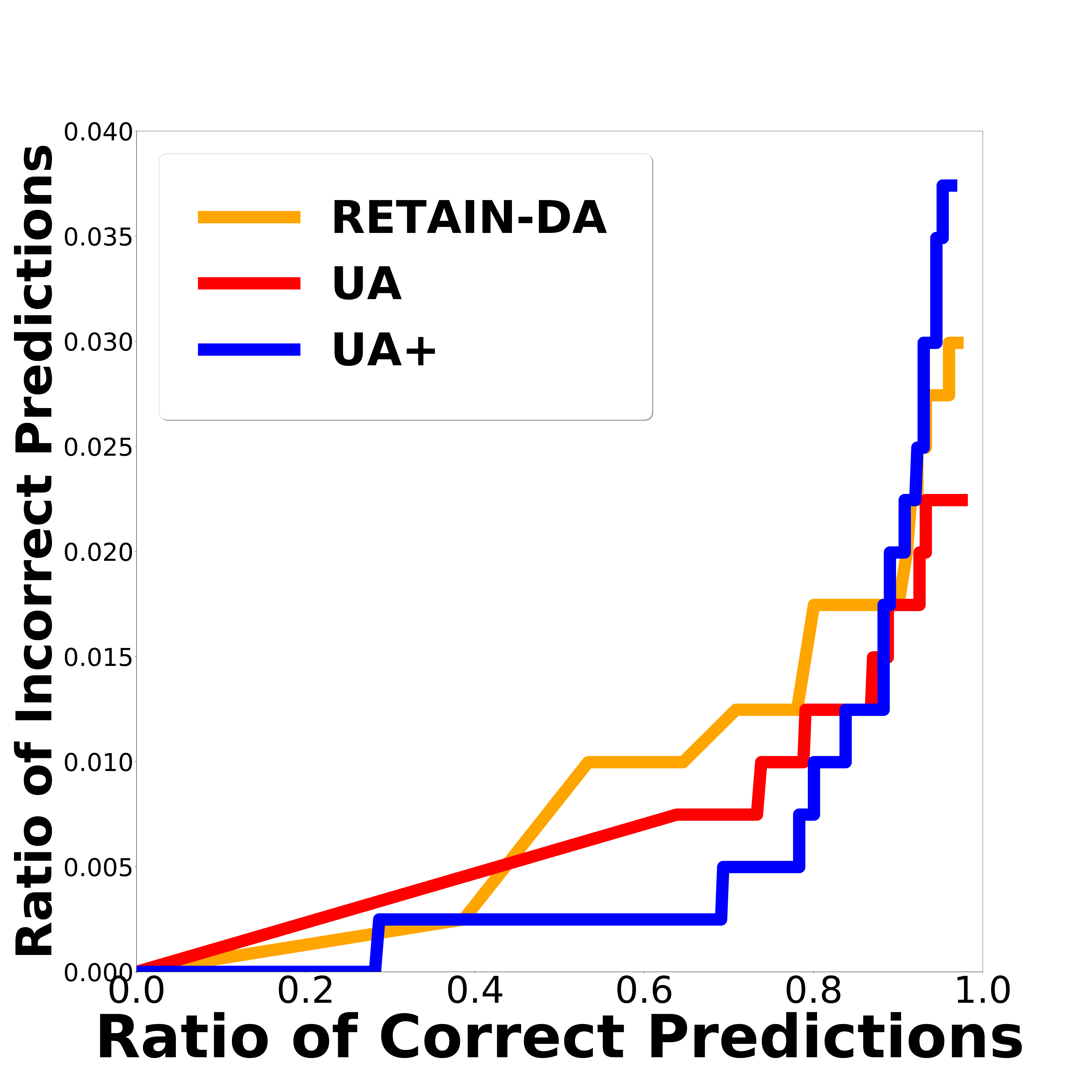}&
\hspace{-0.12in}
\includegraphics[height=2.2cm]{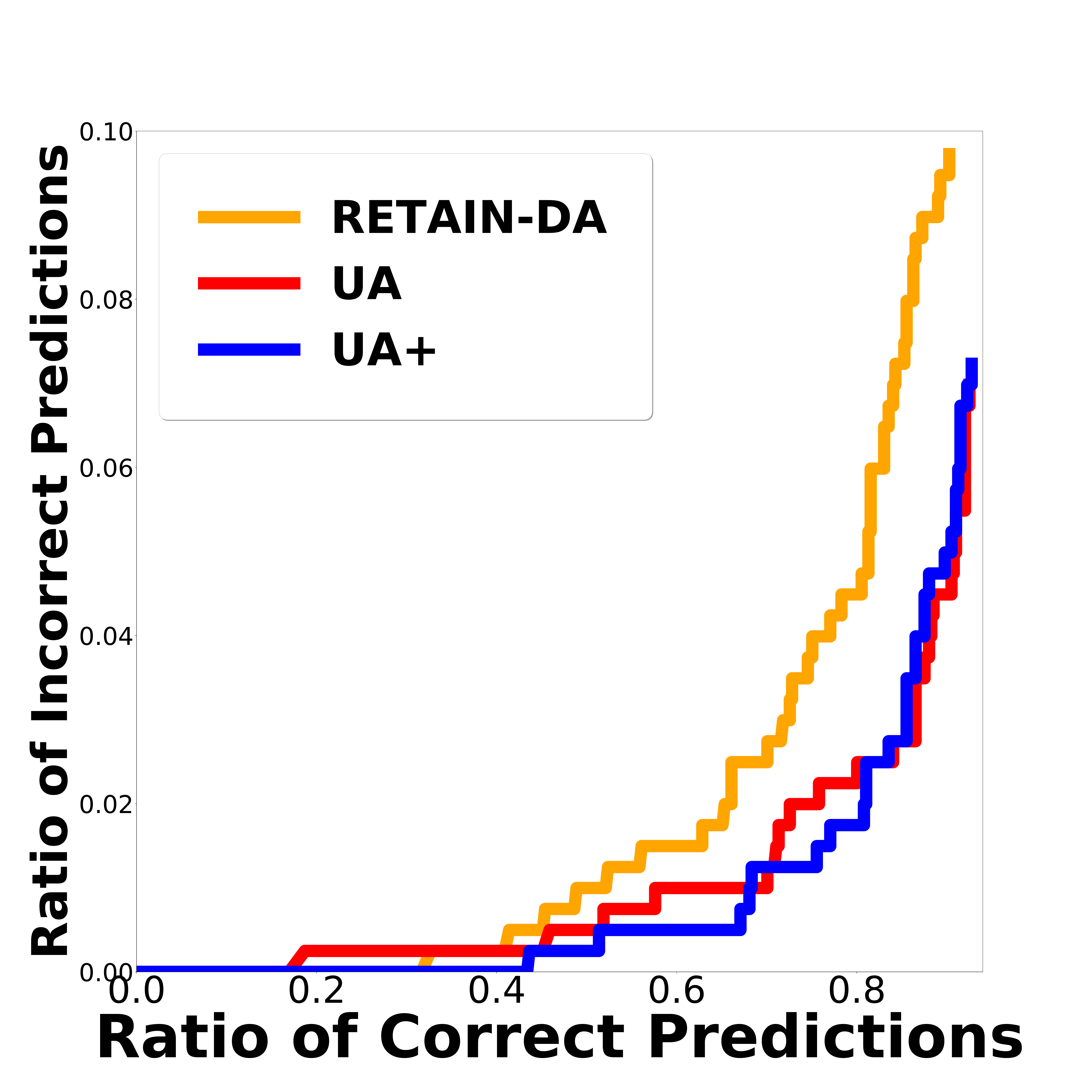}&
\hspace{-0.12in}
\includegraphics[height=2.2cm]{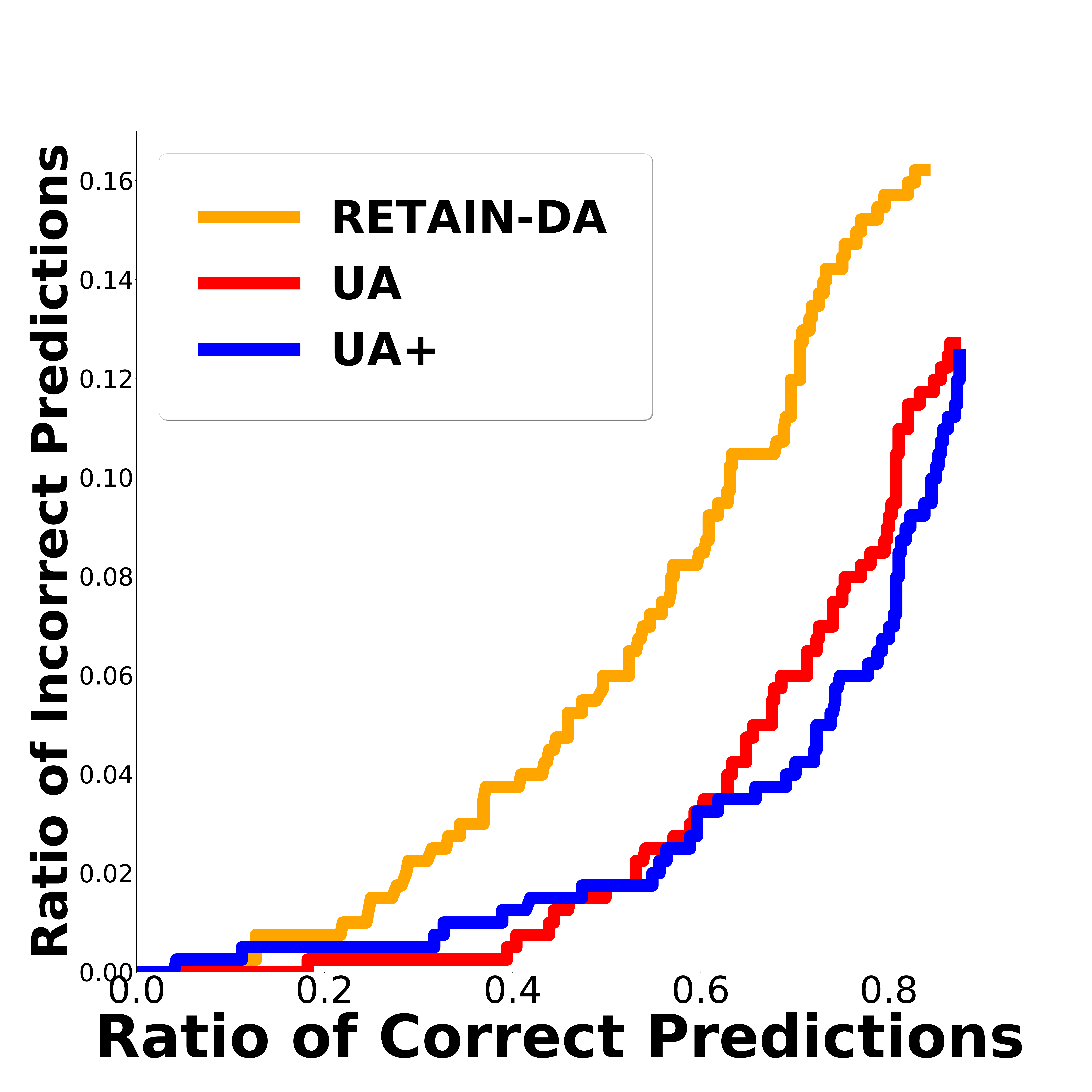}&
\hspace{-0.12in}
\includegraphics[height=2.2cm]{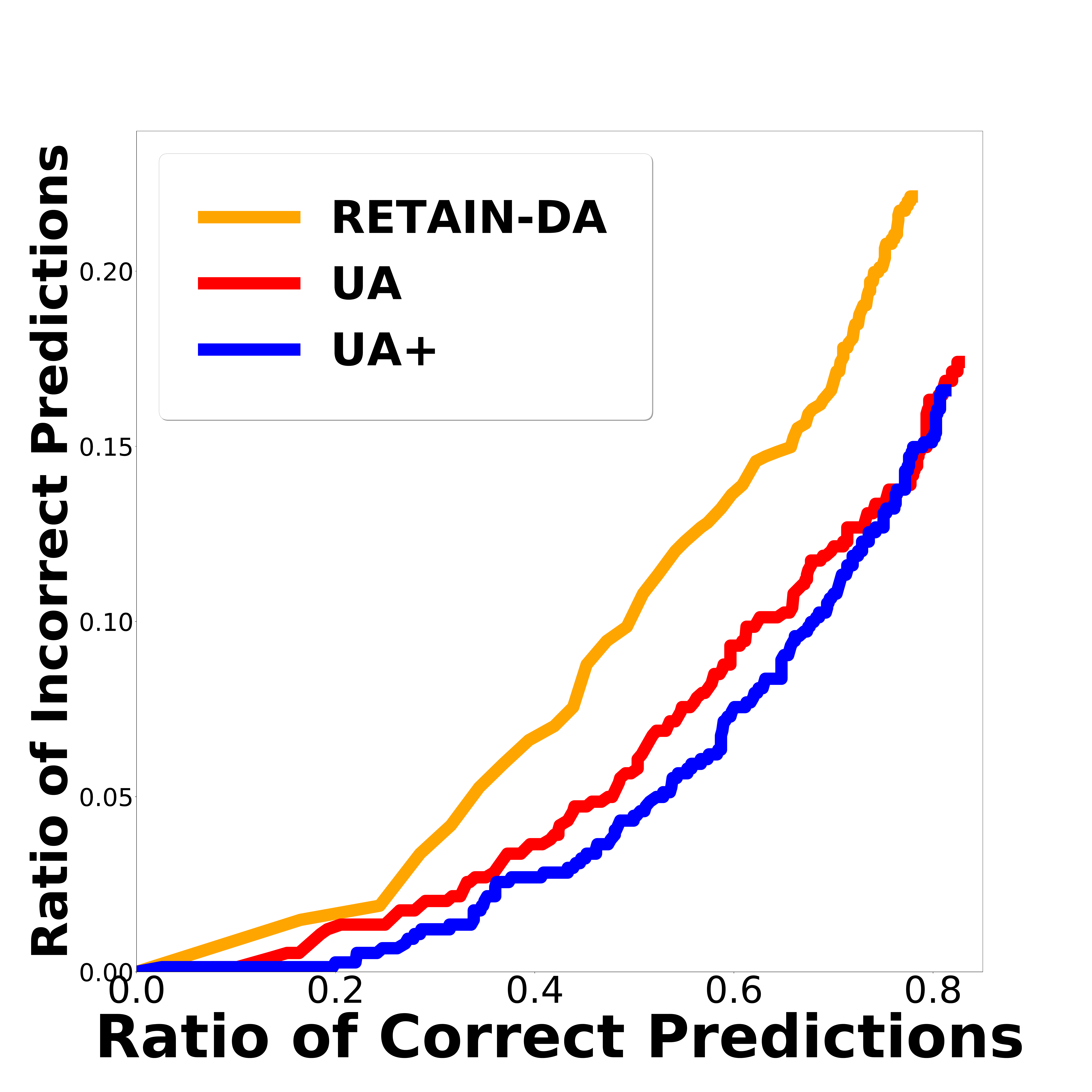}&
\hspace{-0.12in}
\includegraphics[height=2.2cm]{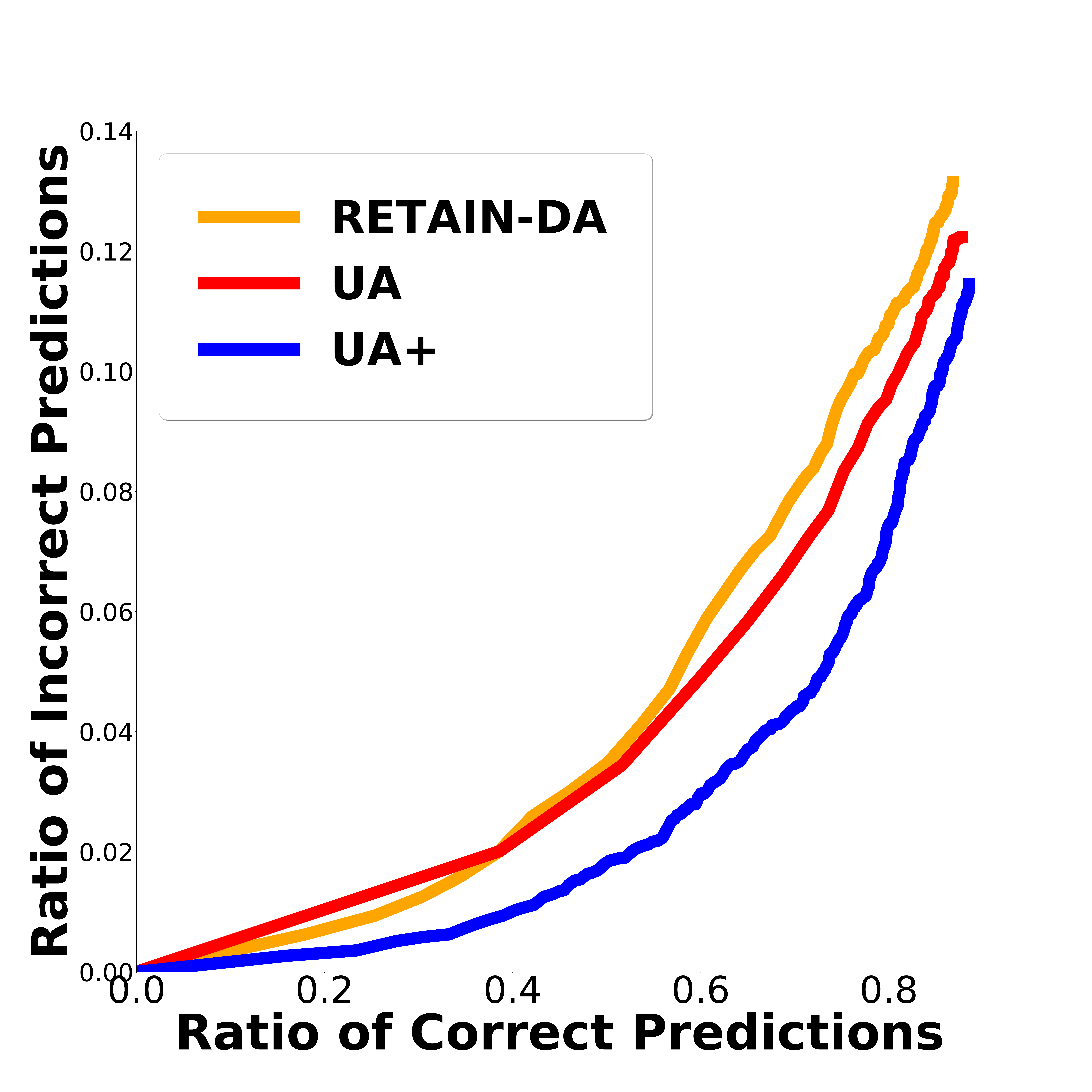}\\
\hspace{-0.10in}(a) PhysioNet & (b) PhysioNet & (c) PhysioNet & (d) PhysioNet & (e) Pancreatic & (f) MIMIC \\
  - Mortality & - Stay $<$ 3 & - Cardiac & - Recovery & Cancer & - Sepsis \\
\end{tabular}
\vspace{-0.16in}  
\caption{\small Experiments on prediction reliability. The line charts show the ratio of incorrect predictions as a function of the ratio of correct predictions for all datasets.}\label{fig:ratio}
\vspace{-0.25in}  
\end{figure*}

\subsection{Evaluation of prediction reliability}
Another important goal that we aimed to achieve with the modeling of uncertainty in the attention is achieving high reliability in prediction. Prediction reliability is orthogonal to prediction accuracy, and~\cite{ece} showed that state-of-the-art deep networks are not reliable as they are not well-calibrated to correlate model confidence with model strength. Thus, to demonstrate the reliability of our uncertainty-aware attention, we evaluate it for the uncertainty calibration performance against baseline attention models in Table~\ref{tbl:ece}, using Expected Calibration Errors (ECE)~\cite{ece} (Eq.~\eqref{eq:ece}). UA and UA+ are significantly better calibrated than RETAIN-DA, RETAIN-SA as well as UA-independent, which shows that independent modeling of variance is essential in obtaining well-calibrated uncertanties.

%For further qualitative analaysis we show the prediction uncertainty of the baseline RETAIN model and our model on all instances from the PhysioNet Challenge in Figure~\ref{fig:uncertainty}. We observe that our model has larger prediction uncertainty for instances whose prediction is incorrect (blue dots above the prediction strength of 0.5 and red dots below 0.5) or weak (near 0.5), while RETAIN shows similar uncertainty for across all inputs. 
%\input{ece_table}

\paragraph{Prediction with ``I don't know" option}
We further evaluate the reliability of our predictive model by allowing it to say I don't know (IDK), where the model can refrain from making a hard decision of yes or no when it is uncertain about its prediction. This ability to defer decision is crucial for predictive tasks in clinical environments, since those deferred patient records could be given a second round examination by human clinicians to ensure safety in its decision. To this end, we measure the uncertainty of each prediction by sampling the variance of the prediction using both MC-dropout and stochastic Gaussian noise over $30$ runs, and simply predict the label for the instances with standard deviation larger than some set threshold as IDK. 

Note that we use RETAIN-DA with MC-Dropout~\cite{rnn_dropout} as our baseline for this experiment, since RETAIN-DA is deterministic and cannot output uncertainty~\footnote{RETAIN-SA is not compared since it largely underperforms all others and is not a meaningful baseline.} We report the performance of RETAIN + DA, UA, and UA+ for all tasks by plotting the ratio of incorrect predictions as a function of the ratio of correct predictions, by varying the threshold on the model confidence (See Figure~\ref{fig:ratio}). We observe that both UA and UA+ output much smaller ratio of incorrect predictions at the same ratio of correct predictions compared to RETAIN + DA, by saying IDK on uncertain inputs. This suggests that our models are relatively more reliable and safer to use when making decisions for prediction tasks where incorrect predictions can lead to fatal consequences. Please see \textbf{supplementary file} for more results and discussions on this experiment. 
\vspace{-0.1in}

\section{Conclusion}
We proposed uncertainty-aware attention mechanism, which generates attention weights following Gaussian distribution with learned mean and variance, that are decoupled and trained in input-adaptive manner. This input-adaptive noise modeling allows to capture heteroscedastic uncertainty, or the instance-specific uncertainty, which in turn yields more accurate calibration of prediction uncertainty. We trained it using variational inference and validated on eight different tasks from three electronic health records, on which it significantly outperformed the baselines and provided more accurate and richer interpretations. Further analysis of prediction reliability shows that our model is accurately calibrated and thus can defer predictions when making prediction with ``I don't know'' option. As future work, we plan to apply our model to tasks such as image annotation and machine translation.

\bibliography{refs}
\bibliographystyle{abbrv}

\title{Supplementary File for Uncertainty-Aware Attention \\ for Reliable Interpretation and Prediction}

%\begin{document}
% \nipsfinalcopy is no longer used
\maketitle

\appendix
\section{Detailed Description of Datasets and Experimental Setup}
\subsection{Datasets}
\paragraph{MIMIC3-Sepsis} We calculated Sepsis-related Organ Failure Assessment Score(SOFA)~\cite{sepsis3} for each patient to determine the onset of sepsis: if SOFA score increases by 2 points or more within the time window, we label the patient as positive. We set the time window as 72 hours, since the current guideline of American Medical Association considers the specified period of suspected infection on sepsis as 48 hours before and up to 24 hours after the onset of sepsis~\cite{sepsis3}. The overal rate of septic patients is 16.07$\%$. Table~\ref{tbl:sepsis_feature_table} describes feature information in details. We selected features under the guidelines of physicians and, for urine outputs, we adopted the similar approach to the recent work~\cite{benchmark_mimic3}: we sum the variables representing urine.

\paragraph{Pancreatic Cancer} This datasets is a subset of electronic healthcare records-based database from healthcare organization, consisting of around 1.5 million records. The database contains demographic information including medical aid beneficiaries, treatmenet information, disease histories, and drug prescription records. In total, 34 features regarding vital signs, social and behavioral factors, medical history, and general information, were extracted from the database over 12 years. Total cholesterol level and fasting glucose levle were sampled after overnight fasting and systolic blood pressure and diastolic blood pressure were checked through medical examinations. Also, there were several questionnaires that are designed to identify social and behavioral risk factors, such as smoking habit, alcohol consumption, and time spent on excercise. Individual medical history was followed with drug perscription history and clinical codes of the 10th revision of the International Classification of Diseases (ICD-10). We determined patients with pancreatic cancer by identifying ICD code, C25, on examination and treatment records. On the labeling process, we exclude those who had previous pancreatic cancer-related treatment records as well as pre-existing medical history of pancreatic cancer. Table~\ref{tbl:pancreas_features} describes feature information in details.

\subsection{Configuration and Parameters} We trained all the models using Adam~\cite{adam} optimizer with dropout regularization. We set the maximum iteration for Adam optimizer as $100,000$, and for other hyperparameters, we searched for the optimal values by cross-validation, within predefined ranges as follows: Mini batch size: $\{32, 64, 128, 256\}$, learning rate: $\{0.01, 0.001, 0.0001\}$, \textit{L-2} regularization: $\{0.02, 0.002, 0.0002, 0.0004\}$, and dropout rate $\{0.1, 0.15, 0.2, 0.25, 0.3, 0.4, 0.5\}$.

\section{Benefits of Input-adaptive Uncertainty Modeling}
We conducted experiments to show the benefits of input-adaptive noise on PhysioNet-Mortality dataset. First, we intentionally corrupted the distribution of original dataset with Gaussian noise. The result shows that UA and UA+ outperform RETAIN in classification performance. Especially, when comparing measured attention weights on noisy features, UA captures 86$\%$ of noisy features, while RETAIN captures only 59$\%$ with a threshold of attention weight, 0.01. For the second experiment, we intentionally increased the original missing rate by 5$\%$, from 92$\%$ to 97$\%$, to simulate low-quality samples. As a result, UA and UA+ models outperform RETAIN in classification performance.

\begin{table*}[h]
\small
\centering
\label{benefit_input_dp}
\begin{tabular}{@{}cccl@{}}
\toprule
\multicolumn{1}{l}{} & Gaussian Noise & 97\% Missing Rate &  \\ \midrule
RETAIN-DA & 0.7692 & 0.7129 &  \\
UA & \textbf{0.7868} & 0.7372 &  \\
UA+ & 0.7864 & \textbf{0.7643} &  \\ \bottomrule
\end{tabular}
\vspace{0.1in}
\caption{\small Classification performance of RETAIN and uncertainty-aware attention models on PhysioNet-Mortality dataset. The reported numbers are AUROC.}
\end{table*}

\begin{figure*}[t]
\vspace{-0.2in}
%\small
\begin{tabular}{c c c}
\hspace{-0.25in}
\includegraphics[height=5cm, width=5.2cm]{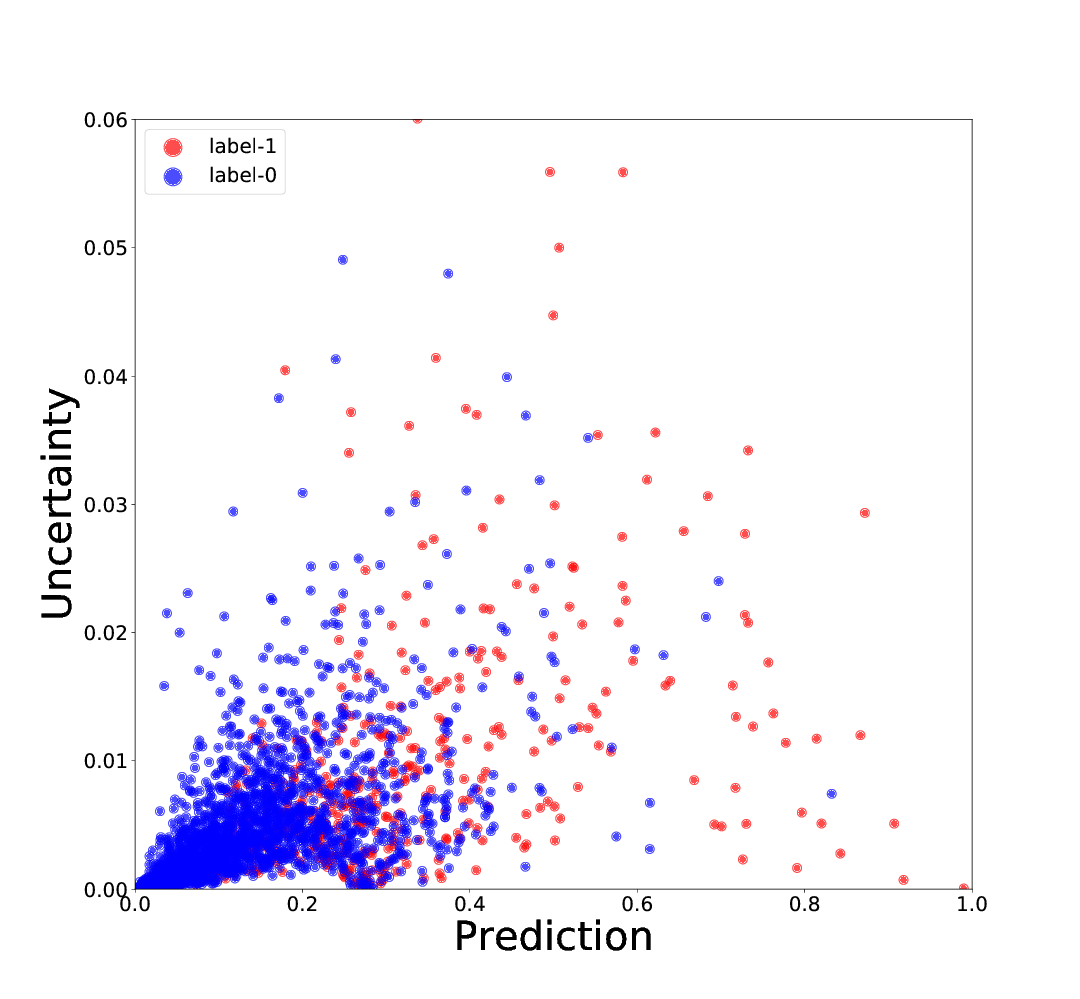}&
\hspace{-0.35in}
\includegraphics[height=5cm, width=5.2cm]{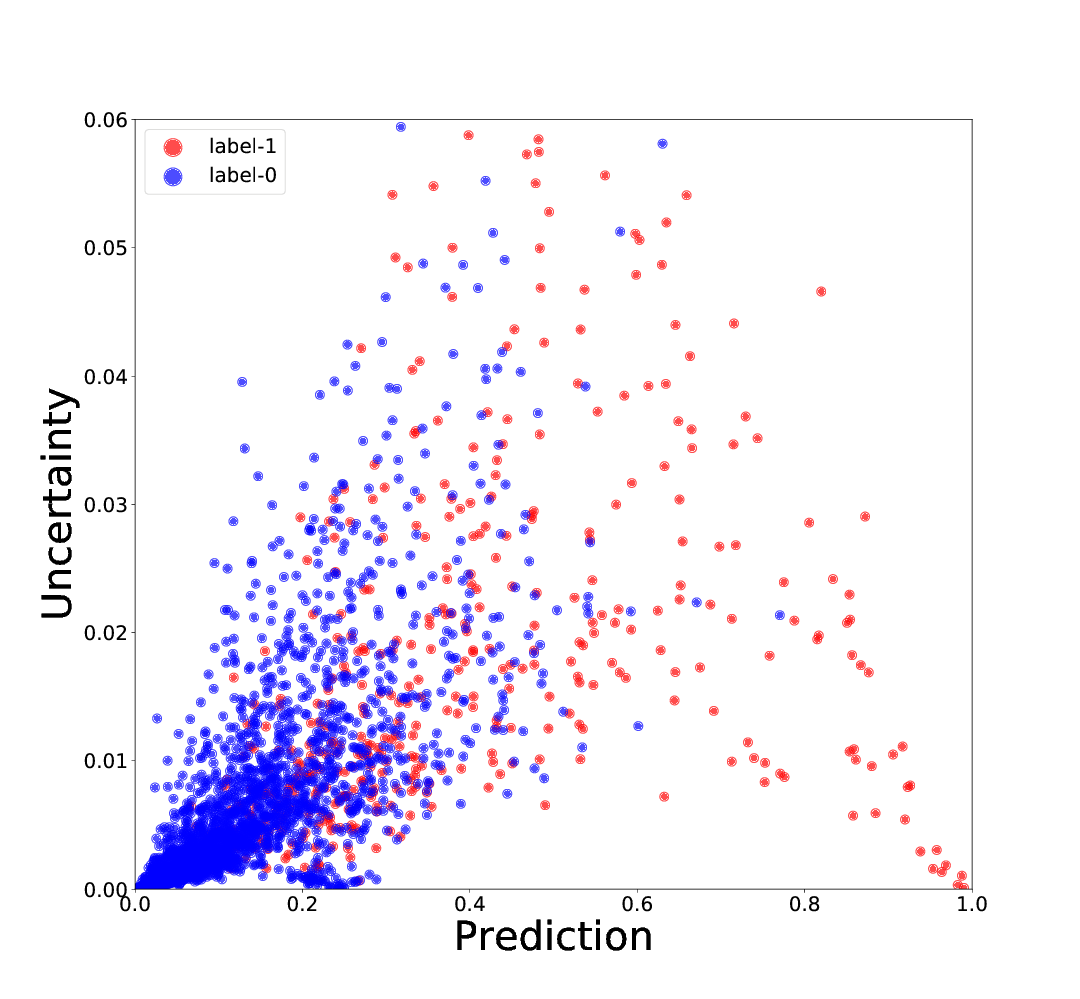}&
\hspace{-0.35in}
\includegraphics[height=5cm, width=5.2cm]{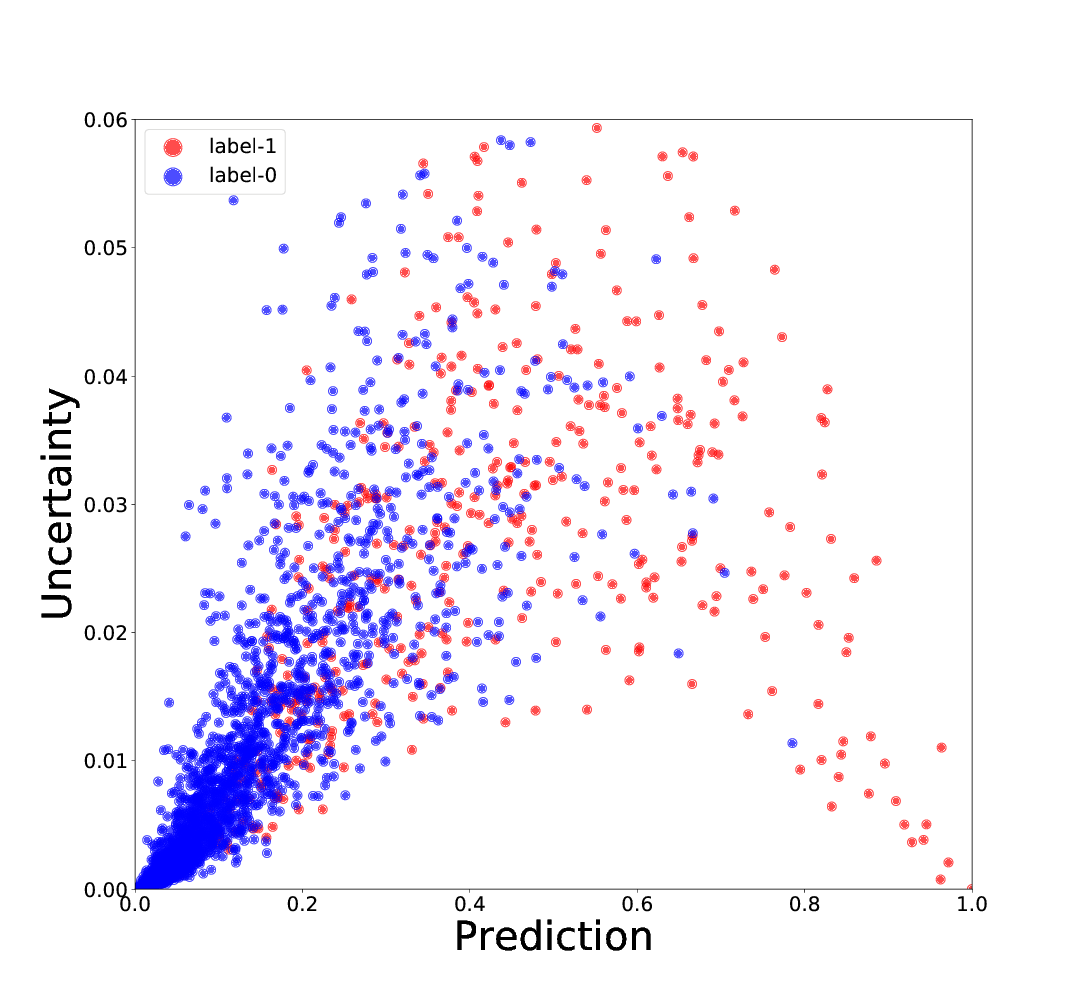}\\
\hspace{-0.30in} (a)RETAIN&(b)UA &(c)UA+ \\
\end{tabular}
\caption{\small Uncertainty over prediction strength on PhysioNet Challenge dataset. For all models, we measured the prediction uncertainty by using MC-dropout with $30$ samples.}\label{fig:uncertainty}
\end{figure*}

\section{Prediction with "I Don't Know" Decision}
We analyzed the predictions for PhysioNet-Mortality to address how many of the IDK predictions would have been false positives, false negatives, or true positives. The result shows that, when correct prediction rate becomes 0.7, UA mainly filters out more false negative cases, while RETAIN filters out more false positive cases. This is a promising result since preventing type II error is critical for healthcare applications.

\begin{table*}[h]
\small
\centering
\label{idk_numbers}
\begin{tabular}{@{}cccc@{}}
\toprule
\multicolumn{1}{l}{} & False Positive & False Negative & True Positive \\ \midrule
RETAIN-DA & \bf 14  & 14     &  8  \\
%\hline
UA        &  7      & \bf 22 & \bf 10  \\
%\hline
UA+       &  8      & 21     &  9  \\ \bottomrule
\end{tabular}
\vspace{0.1in}
\caption{\small Number of false positives, false negatives, and true positives in IDK holder on PhysioNet-Mortality dataset.}
\end{table*}

\begin{figure*}[t]
\small
\begin{tabular}{c c c}
\hspace{0.05in}
\includegraphics[height=3cm, width=4cm]{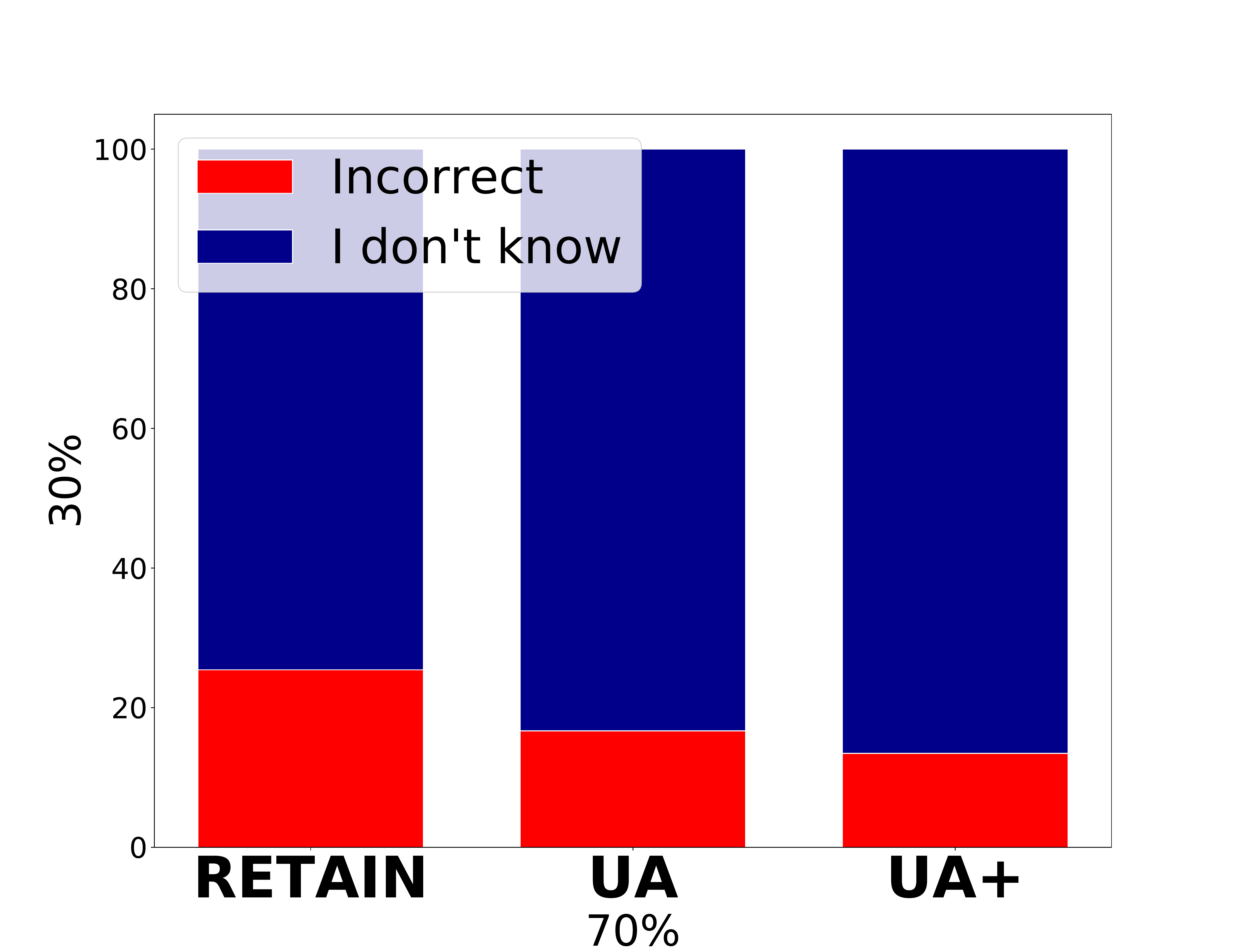}&
\hspace{0.07in}
\includegraphics[height=3cm, width=4cm]{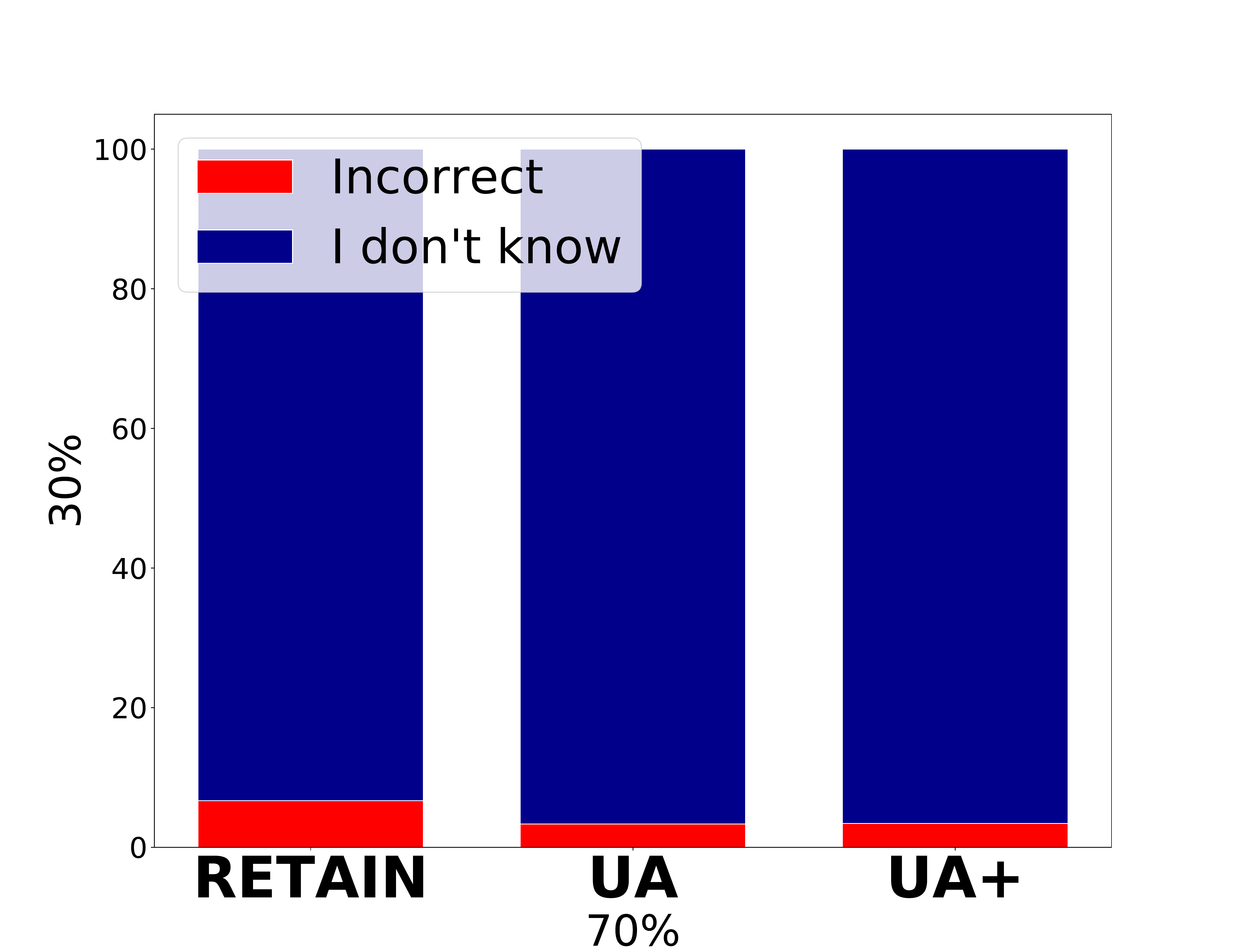}&
\hspace{0.07in}
\includegraphics[height=3cm, width=4cm]{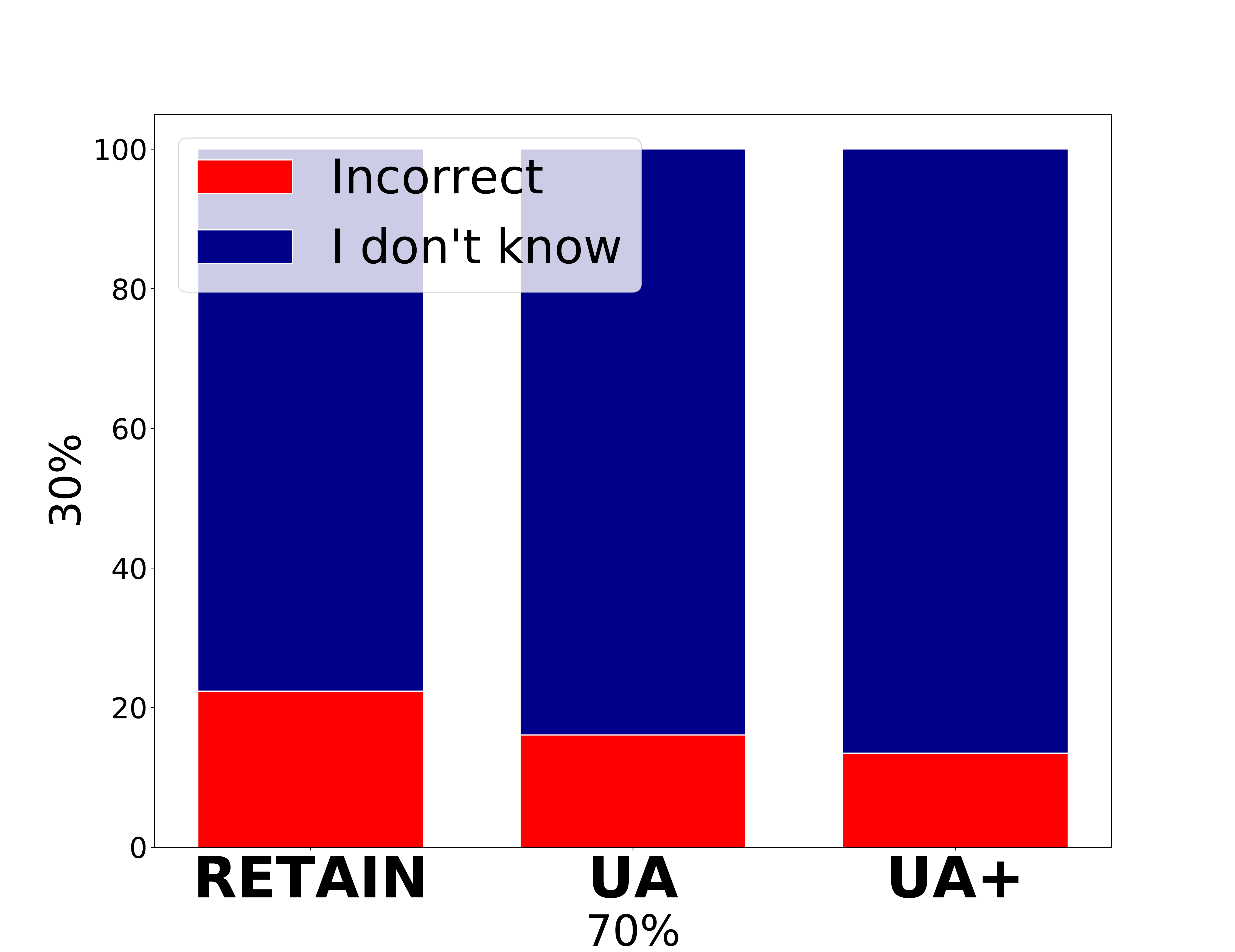}\\
\hspace{0.04in}(a) PhysioNet-Mortality & (b) PhysioNet-Stay & (c) PhysioNet-Cardiac \\

\end{tabular}  
\begin{tabular}{c c c}
\hspace{0.05in}
\includegraphics[height=3cm, width=4cm]{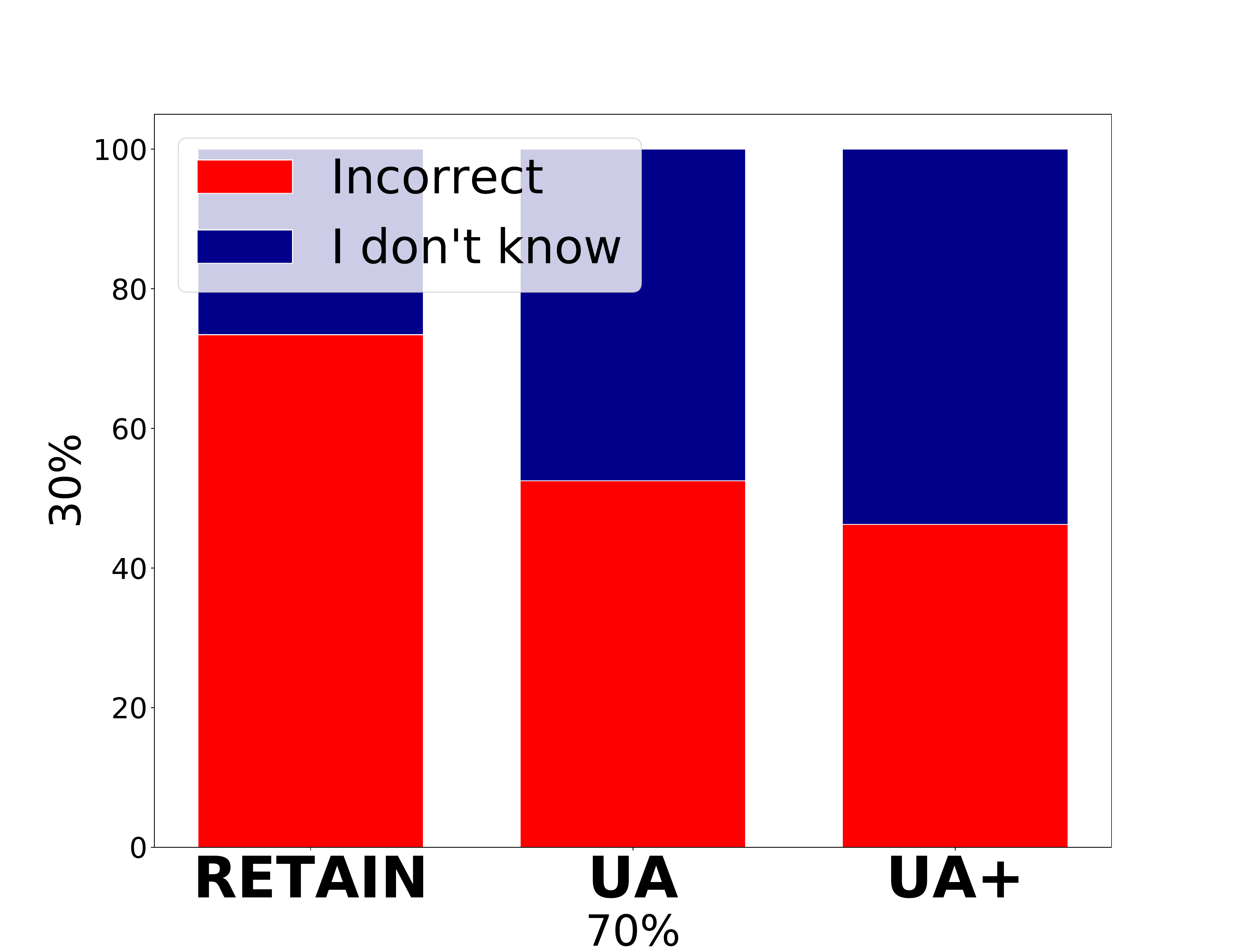}&
\hspace{0.07in}
\includegraphics[height=3cm, width=4cm]{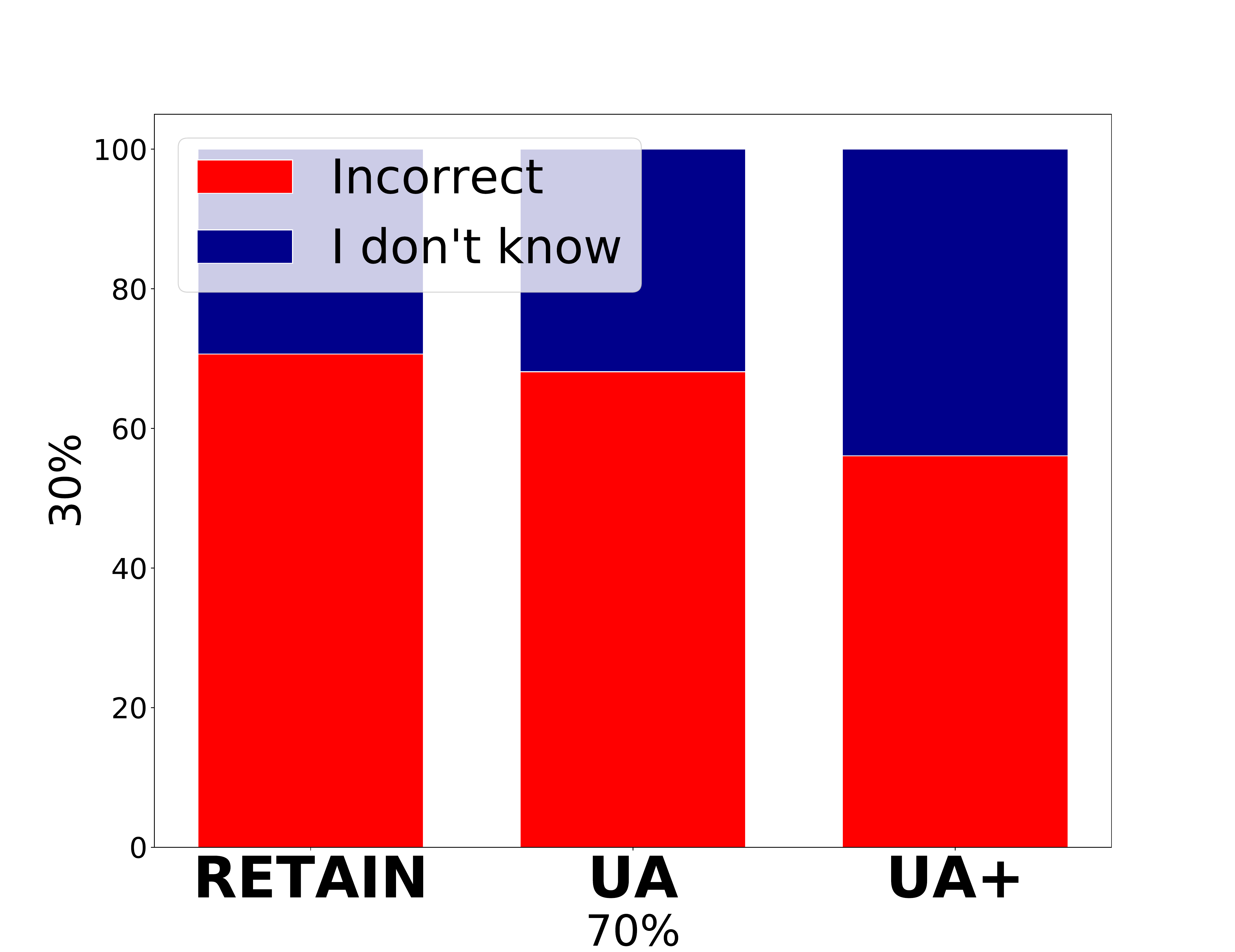}&
\hspace{0.07in}
\includegraphics[height=3cm, width=4cm]{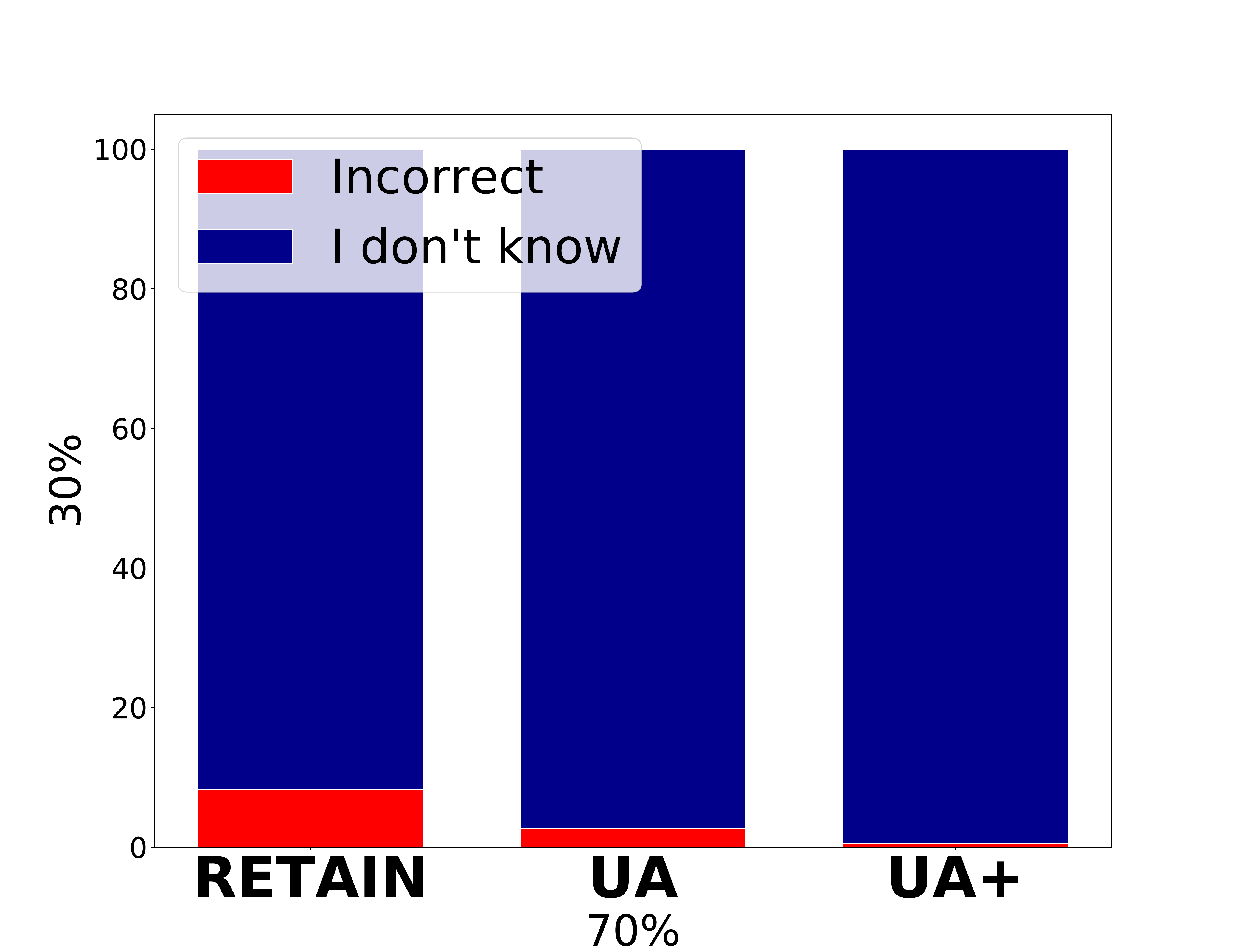}\\
\hspace{0.04in} (d) PhysioNet-Recovery & (e) Pancreatic Cancer & (f) MIMIC-Sepsis \\
\end{tabular}
\vspace{-0.1in}  
\caption{\small Experiments on prediction reliability. The stacked bar charts show the ratio of IDK and incorrect predictions, when correct prediction becomes 0.7.}\label{fig:supple_idk}
\end{figure*}

In Figure~\ref{fig:supple_idk}, we observe that both UA and UA+ are more likely to say IDK rather than make incorrect predictions when compared against RETAIN + MC Dropout model, which suggests that they are relatively more reliable, and safer to use for making clinical decisions where incorrect predictions can lead to fatal consequences. For instance, on sepsis prediction task, UA+ made incorrect prediction only on $0.17\%$ of the instances ($0.80\%$ for UA) while avoiding $29.83\%$ of potentially incorrect predictions based on uncertainty, when correct prediction rate becomes 0.7. On the other hand, RETAIN + MC Dropout predicted incorrectly on $2.51\%$ of the instances with $27.68\%$ IDK predictions. Considering the consequences that follow an incorrect prediction of sepsis, this is a significant difference. Furthermore, for pancreatic cancer prediction task, our model made $14.32\%$ incorrect predictions with $15.68\%$ IDK decisions, while RETAIN + MC Dropout made incorrect prediction on $17.54\%$ of instances with $12.46\%$ IDK decisions. This difference is significant considering the severe consequences an incorrect cancer prediction has on the patient.

\begin{table*}[t]
\small
\vspace{-0.3in}
\begin{center}
\label{tbl:sepsis_feature_table}
\caption{Feature information of MIMIC-Sepsis dataset.}
\vspace{0.1in}
\begin{tabular}{|l|l|l|}
\hline
\rowcolor[HTML]{EFEFEF} 
\multicolumn{1}{|c|}{\cellcolor[HTML]{EFEFEF}\textbf{Features}} & \multicolumn{1}{c|}{\cellcolor[HTML]{EFEFEF}\textbf{Item-ID}}                       & \multicolumn{1}{c|}{\cellcolor[HTML]{EFEFEF}\textbf{Name of Item}}                                                                                                                                                              \\ \hline
Age                                                             & N/A                                                                                 & \begin{tabular}[c]{@{}l@{}}intime\\ dob\end{tabular}                                                                                                                                                                            \\ \hline
Heart rate                                                      & \begin{tabular}[c]{@{}l@{}}211\\ 22045\end{tabular}                                 & \begin{tabular}[c]{@{}l@{}}Heart Rate\\ Heart Rate\end{tabular}                                                                                                                                                                 \\ \hline
FiO2                                                            & \begin{tabular}[c]{@{}l@{}}223835\\ 3420\\ 3422\\ 190\end{tabular}                  & \begin{tabular}[c]{@{}l@{}}Inspired O2 Fraction\\ FiO2\\ FiO2 {[}Meas{]}\\ FiO2 set\end{tabular}                                                                                                                                \\ \hline
Temperature                                                     & \begin{tabular}[c]{@{}l@{}}676\\ 678\\ 223761\\ 223762\end{tabular}                 & \begin{tabular}[c]{@{}l@{}}Temperature C\\ Temperature F\\ Temperature Fahrenheit\\ Temperature Celsius\end{tabular}                                                                                                            \\ \hline
Systolic Blood Pressure                                         & \begin{tabular}[c]{@{}l@{}}51\\ 442\\ 455\\ 6701\\ 220179\\ 220050\end{tabular}     & \begin{tabular}[c]{@{}l@{}}Arterial BP{[}Systolic{]}\\ Manual BP{[}Systolic{]}\\ NBP{[}Systolic{]}\\ Arterial BP \#2 {[}Systolic{]}\\ Non Invasive Blood Pressure{[}systolic{]}\\ Arterial Blood Pressure{[}systolic{]}\end{tabular}   \\ \hline
Diastolic Blood Pressure                                        & \begin{tabular}[c]{@{}l@{}}8368\\ 8440\\ 8441\\ 8555\\ 220051\\ 220180\end{tabular} & \begin{tabular}[c]{@{}l@{}}Arterial BP{[}Diastolic{]}\\ Manual BP{[}Diastolic{]}\\ NBP{[}Diastolic{]}\\ Arterial BP \#2{[}Diastolic{]}\\ Non Invasive Blood Pressure{[}Diastolic{]}\\ Arterial Blood Pressure{[}Diastolic{]}\end{tabular} \\ \hline
PaO2                                                            & \begin{tabular}[c]{@{}l@{}}50821\\ 50816\end{tabular}                               & \begin{tabular}[c]{@{}l@{}}PO2\\ Oxygen\end{tabular}                                                                                                                                                                            \\ \hline
GCS - Verbal Response                                           & 223900                                                                              & Verbal Response                                                                                                                                                                                                                 \\ \hline
GCS - Motor Response                                            & 223901                                                                              & Motor Response                                                                                                                                                                                                                  \\ \hline
GCS - Eye Opening                                               & 220739                                                                              & Eye Opening                                                                                                                                                                                                                     \\ \hline
Serum Urea Nitrogen Level                                       & 51006                                                                               & Urea Nitrogen                                                                                                                                                                                                                   \\ \hline
Sodium Level                                                    & 950824                                                                              & Sodium Whole Blood                                                                                                                                                                                                              \\ \hline
White Blood Cells Count                                         & \begin{tabular}[c]{@{}l@{}}51300\\ 51301\end{tabular}                               & \begin{tabular}[c]{@{}l@{}}WBC Count\\ White Blood Cells\end{tabular}                                                                                                                                                           \\ \hline
Urine Output                                                    & \begin{tabular}[c]{@{}l@{}}40055\\ 43175 \\ 40069 \\ 40094 \\ 40715 \\40473 \\ 40085 \\ 40057 \\ 40056 \\ 40405 \\ 40428 \\ 40086 \\ 40096 \\ 40651 \\ 226559 \\ 226560 \\ 226561 \\ 226584 \\ 226563 \\ 226564 \\ 226565 \\ 226567 \\ 226557 \\ 226558 \\ 227488 \\ 227489 \end{tabular}                                        & \begin{tabular}[c]{@{}l@{}}Urine Out Foley \\ Urine \\ Urine Out Void \\ Urine Out Condom Cath \\ Urine Out Suprapubic \\ Urine Out IleoConduit \\ Urine Out Incontinent \\ Urine Out Rt Nephrostomy \\ Urine Out Lt Nephrostomy \\ Urine Out Other \\ Urine Out Straight Cath \\ Orine Out Incontinent \\ Urine Out Ureteral Stent $\#$1 \\ Urine Out Ureteral Stent $\#$2 \\ Foley \\ Void \\ Condom Cath \\ Ileoconduit \\ Suprapubic \\ R Nephrostomy \\ L Nephrostomy \\ Straight Cath \\ R Ureteral Stent \\ L Ureteral Stent \\ GU Irrigant Volumne In \\ GU Irrigant/Urine Volume Out\end{tabular} \\ \hline
\end{tabular}
\end{center}
\end{table*}

% Please add the following required packages to your document preamble:
% \usepackage[table,xcdraw]{xcolor}
% If you use beamer only pass "xcolor=table" option, i.e. \documentclass[xcolor=table]{beamer}
\begin{table}[h]
\small
\centering
\caption{Feature information of pancreatic cancer dataset.}
\label{tbl:pancreas_features}
\begin{tabular}{|l|l|}
\hline
\rowcolor[HTML]{EFEFEF} 
\multicolumn{1}{|c|}{\cellcolor[HTML]{EFEFEF}{\color[HTML]{000000} \textbf{Category}}} & \multicolumn{1}{c|}{\cellcolor[HTML]{EFEFEF}{\color[HTML]{000000} \textbf{Feature}}} \\ \hline
Demographics & \begin{tabular}[c]{@{}l@{}}Age\\ Sex\end{tabular} \\ \hline
Socio-Economic Status & \begin{tabular}[c]{@{}l@{}}Income Level\\ Type of Disability \end{tabular} \\ \hline
Health Screening & \begin{tabular}[c]{@{}l@{}}Body Mass Index (BMI)\\ Waist Circumference\\ Systolic Blood Pressure\\ Diastolic Blood Pressure\\ Fasting Glucose\\ Total Cholesterol\\ Triglyceride\\ Hemoglobin\\ Urine Protein\\ Creatinine\\ HDL Cholesterol\\ LDL Cholesterol\\ Aspartate Aminotransferase\\ Alanine Transaminase\\ Gamma-Glutamyl Transferase\end{tabular} \\ \hline
Family History & \begin{tabular}[c]{@{}l@{}}Liver Disease\\ Stroke\\ Heart Disease\\ Hypertension\\ Diabetes Mellitus\\ Cancer\end{tabular} \\ \hline
Personal History & \begin{tabular}[c]{@{}l@{}}Stroke or Cerebral Infarction-related Disease\\ Heart Disease\\ Hypertension\\ Diabetes Mellitus\\ Hyperlipidemia\\ Tuberculosis\end{tabular} \\ \hline
Social and behavioral Factor & \begin{tabular}[c]{@{}l@{}}Alcohol Consumption\\ Smoking Habit\\ Physical Exercise\end{tabular} \\ \hline
\end{tabular}
\end{table}

%\bibliographystyle{abbrv}
%\bibliography{refs}

\end{document}